%% file: p0-mct-main.tex
\documentclass[preprint,12pt]{elsarticle}
\usepackage[utf8]{inputenc} 
\usepackage[T1]{fontenc}    
\usepackage{url}            
\usepackage{booktabs}       
\usepackage{amsfonts}       
\usepackage{nicefrac}       
\usepackage{microtype}      
\usepackage{lipsum}
\usepackage{times}
\usepackage{amsmath}
\usepackage{graphicx}
\usepackage{multicol}
\usepackage{longtable}
\usepackage[refpages]{gloss}
\usepackage{float}
\usepackage{anysize}
\usepackage{appendix}
\usepackage{lscape} 
\usepackage{pdflscape}
\usepackage{multirow}
\usepackage{listings}
\usepackage{color}
\usepackage{setspace}
\usepackage{enumerate} 
\usepackage{natbib}
\usepackage{algpseudocode}
\usepackage{algorithm}
\usepackage{rotating}
\usepackage{multirow}
\usepackage{array}
\usepackage{graphicx}
\usepackage{amssymb}
\usepackage{fixltx2e}
\usepackage{emojione}
\usepackage{CJKutf8}
\usepackage{scrextend}
\usepackage[colorlinks]{hyperref}
\usepackage{graphicx}
\usepackage{amssymb}
\begin{document}

\begin{frontmatter}


\title{A multilevel clustering technique for community detection}


\author[label1]{Isa Inuwa-Dutse}
\address[label1]{School of Computer Science\\
        University of St Andrews}
%\address[label2]{University of Hertfordshire}

\author[label2]{Mark Liptrott}
\address[label2]{Department of Computer Science\\
 Edge Hill University, UK}

\author[label2]{Yannis Korkontzelos}
%\address[label2]{Department of Computer Science\\
% Edge Hill University, UK}

%\address{California, United States}
%\author{Author names concealed for blind review}
%\author{
%  Isa Inuwa-Dutse\thanks{Corresponding author: isadutse91@gmail.com} 
% Mark Liptrott 
% Yannis Korkontzelos \\
% Department of Computer Science\\
% Edge Hill University, UK }

\begin{abstract} 
A network is a composition of many communities, i.e., sets of nodes and edges with stronger relationships, with distinct and overlapping properties. 
Community detection is crucial for various reasons, such as serving as a functional unit of a network that captures local interactions among nodes. 
Communities come in various forms and types, ranging from biologically to technology-induced ones. 
As technology-induced communities, social media networks such as Twitter and Facebook connect a myriad of diverse users, leading to a highly connected and dynamic ecosystem.  
Although many algorithms have been proposed for detecting socially cohesive communities on Twitter, mining and related tasks remain challenging.
This study presents a novel detection method based on a scalable framework to identify related communities in a network. 
We propose a multilevel clustering technique \textit{(MCT)} that leverages structural and textual information to identify local communities termed \textit{microcosms}. 
Experimental evaluation on benchmark models and datasets demonstrate the efficacy of the approach. 
This study contributes a new dimension for the detection of cohesive communities in social networks. 
The approach offers a better understanding and clarity toward describing how low-level communities evolve and behave on Twitter. 
From an application point of view, identifying such communities can better inform recommendation, among other benefits. 
\end{abstract}


\end{frontmatter}

%\linenumbers

% INTRODUCTION:
\input{p1-introduction.tex}

% BACKGROUND: ... and related work
\input{p2-background.tex}

% MICROCOSM DETECTION 
\input{p3-microcosm-detection.tex}

% EXPERIMENTATION/DISCUSSION
\input{p4-experimentation.tex}

% CONCLUSION
\input{p5-conclusion.tex}

\section*{Acknowledgements} 
%The authors would like to thank \textit{[name concealed for blind review]} for the fruitful discussions and exchange of ideas about a multitude of aspects related to social media, spam content and the motives of spammers. 
The third author has participated in this research work as part of the \textit{[project name concealed for blind review]}
%\textit{TYPHON} 
Project, which has received funding from the European Union’s Horizon 2020 Research and Innovation Programme under grant agreement \textit{[agreement number concealed for blind review]}.
%No.~780251.

%%%APPENDIX
%\newpage
\input{pn-appendix.tex}
\bibliographystyle{elsarticle-num-names}
\bibliography{pn-mct-references}

\end{document}

%% file: p1-introduction.tex
\section{Introduction}
\label{sec:introduction} 

A network comprises of many sub-networks or communities with distinct and overlapping properties. 
Networks exhibit varying degrees of organisations \cite{lancichinetti2009detecting}, and discovering the structure of various network forms has been investigated \cite{scott1988social,watts1998collective,albert2002statistical,newman2004finding}. 
As network size increases, so does the possibility of fragmentation \cite{berelson1964human,shaw1971group}, leading to a decrease in the homogeneity of behaviour and attitude across groups \cite{granovetter1992problems}. 
Because similarity breeds attraction and interaction \cite{brass1998relationships}, network communities are defined by sets of nodes and edges with strong relationships. % which are expressed as a function of relatedness. The set of nodes with a high degree of relationship constitutes a local 
Communities are a fundamental organisation principle, especially in vast networks, allowing to analyse the structure and function of networks \cite{watts1998collective,newman2003social}. 
Identifying local network structures: 
(a) provides a means for complex network analysis \cite{williams2000simple}, for applications such as the detection of inter-related web-pages \cite{flake2002self,papadopoulos2012community},
(b) allows to detect cliques \cite{newman2003social} and facilitates intelligent recommendations \cite{cao2015improved} 
(c) allows to discover organisational principles of networks \cite{newman2003properties,yang2013community}, and 
(d) helps in studying social behaviour of users \cite{arnaboldi2013egocentric}. Examples of biological, social or technological networks where community detection has been applied are:
    \textit{protein--protein interaction networks} \cite{krogan2006global}, 
    \textit{social networks} \cite{albert2002statistical,newman2003social}, 
    \textit{food webs} \cite{williams2000simple},
    \textit{collaboration networks}, \cite{nascimento2003analysis} and
    the \textit{World Wide Web} \cite{albert2002statistical}. 

The underlying difference across many network communities refers to the definition of connections: some are deterministic, while some are just probabilistic and potentially non-deterministic. %Of the technologically-induced communities, 
Social media, e.g., \textit{Twitter} %\footnote{\url{https://twitter.com}} 
and \textit{Facebook}, %\footnote{\url{https://facebook.com}} 
connect a myriad of diverse users, leading to a highly connected, dynamic ecosystem. 
The complexity and dynamism of this ecosystem results in multiple  interaction types at various layers of granularity and intensity: global or local, positive or negative, influential or not, high or low-level. 
Such interactions culminate in the formation of communities at various levels. 
Despite the proliferation of various community detection methods, identifying socially cohesive communities on Twitter still poses challenges. 
Communities with low presence are implicit and require extensive exploration to understand the mechanism governing their behaviour \cite{palla2007quantifying}. 
Since social networks exhibit properties from other networks \cite{newman2003social}, the limitations of existing approaches are due to:

\begin{figure}[t]
    \centering
    \includegraphics[width=\linewidth]{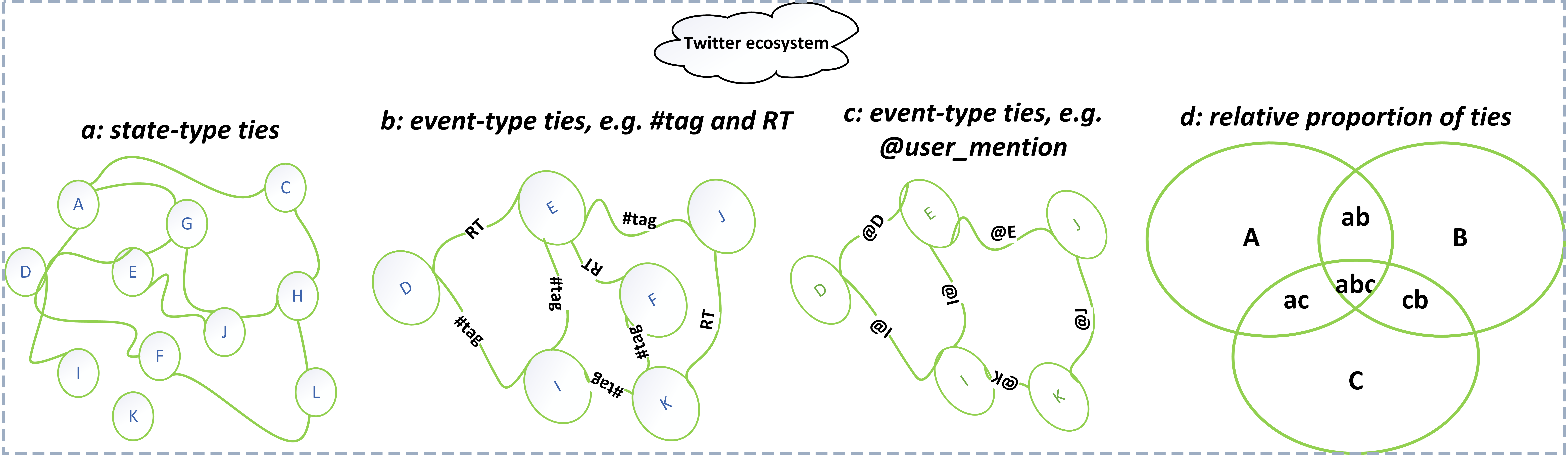}
    \caption[Topological structure on Twitter]{Examples of \textit{event-type ties} ((a),(b) and (c)) on Twitter, allowing users to openly connect via: (a) unidirectional or directed means (e.g.~friend or follower), bidirectional or undirected (among friends and followers) (b -- c), and indirect or \textit{transitory events}: \textit{retweets, mentions} or \textit{likes}. 
    These flexible connections challenge cohesive community detection and contribute to the proliferation of spurious content. In (d), $A = \{a,a_1,...,a_l\},B=\{b,b_1,...,b_m\}, \mbox{ and } C=\{c,c_1,...,c_n\}$ denote users. 
    Reciprocal ties (e.g.~$ac$) or transitive ties (e.g.~$abc$) are rare.}
    \label{fig:Twitter-ecosystem}
\end{figure}

\paragraph{Methodological viewpoints and connection types} 
Social network theorists hold two viewpoints in investigating social relationships in a network: realist, based on a pre-conceived notion of the existence of relationships, and nominalist, based on questions posed by the investigator \cite{laumann1989boundary}. 
Moreover, social ties are formed around \textit{event-type ties}, a transitory connection that often results in socially distant members. 
Such connections on Twitter include subscribing to trendy hashtags or retweeting popular users. 
\textit{State-type ties} are based on static, structural connections among users, which suggest familiarity and trust \cite{borgatti2011network}. 
Community detection on Twitter focusses mostly on directed connections (\textit{event-type ties}) based on the \textit{realist's} approach. 
This is valid in many networks, but can lead to many unrelated sets of users.
We argue that the wealth of connection forms on Twitter, shown in Figure~\ref{fig:Twitter-ecosystem}, contribute to widespread spurious content and imply the existence of less cohesive user communities. 
%Establishing the equivalence of network entities is achieved either based on (1) equivalent units with the same connection pattern to the same neighbours or (2) equivalent units have the same or a similar connection pattern to different neighbours \cite{doreian2005positional}. Accordingly, communities are further formed around two primary modalities or sources of information: \textit{network structure} and \textit{attributes of nodes}. However, existing studies mostly focus on one aspect; the few studies based on a bi-modal source are limited in addressing the nuances on Twitter.
        
\paragraph{Proliferation and complexity of online content} a rapid increase in network size increases the likelihood of fragmentation \cite{berelson1964human,shaw1971group}, which in turn decreases the homogeneity of behaviour and attitude across groups \cite{granovetter1992problems}. 
With an average 139m daily users contributing to 500m content\footnote{See \url{www.omnicoreagency.com/twitter-statistics}}, it is becoming more challenging to keep track of socially cohesive communities on Twitter. 
Furthermore, large scale and transitory content (mostly from influential users) often dominate the space leading to many forms of explicit communities \cite{kwak2010twitter}. 
Thus, basing a community detection task on transitory aspects of \textit{metadata} such as popular hashtags or trending topics does not often reflect true connectivity \cite{wilson2009user}, hence limiting the full realisation of the benefits in communities such as \textit{cliquishness} and \textit{local coherence}. 

\noindent This study attempts to address the identified challenges, to advance our knowledge concerning community detection problems. 

\subsection{A Multilevel Clustering Technique}
\label{sec:introducing-mct}

A community detection paradigm involves prediction and quantification to identify a community structure and relevant details about a network \cite{chen2009detecting}. 
Predicting membership and assigning items to clusters is achieved using equivalence measures or scoring functions. 
Establishing the equivalence of network entities is achieved based on 
(a) equivalent units with the same connection pattern to the same neighbours or 
(b) equivalent units that have the same or a similar connection pattern to different neighbours \cite{doreian2005positional}. 
Accordingly, communities are formed around two primary modalities or information sources: \textit{network structure} and \textit{node attributes}. 
Until recently, community detection methods relied on a single information source. 
Conventional methods such as \textit{normalised cut} \cite{shi2000normalized} and modularity \cite{newman2006modularity} rely on the topological structure of networks. 
A bi-modal approach, based on network structure and the corresponding features or attributes of nodes as information sources, is becoming popular \cite{balasubramanyan2011block,lin2012community,leskovec2012learning,yang2013community}. %However, existing studies mostly focus on one aspect; the few studies based on a bi-modal source are limited in addressing the nuances on Twitter.
According to Figure~\ref{fig:Twitter-ecosystem}, connections on \textit{Twitter} may manifest differently, such as \textit{sharing a link}, \textit{re-tweeting (RT)}, using the same or similar \textit{hashtags}, \textit{user mention (@)} or \textit{follower-ship}. 
Such connections are porous, allowing to connect with many diverse users and hindering the identification of cohesive groups. 
%The problem of community detection on Twitter is mostly centred around a directed form of connections, i.e.~\textit{event-type ties} and adopts the \textit{realist's} approach. While this is valid in many networks, such an approach could lead to many unrelated sets of users. However, by focusing on smaller groups with a high degree of \textit{structural} and \textit{textual} similarities, it is possible to identify communities, which are homogeneous to many sociodemographic behavioural, and intrapersonal characteristics \cite{miller2001birds}.
Noting that these eccentric connections patterns can mislead the detection of socially related users and encourage the propagation of spurious content, we propose a \textit{multilevel clustering technique (MCT)} to identify socially cohesive user groups on Twitter, termed \textit{microcosms}. 
No practical reasons prevent MCT to apply to other domains that involve network data. 
However, it would require minor amendments for platforms where a reciprocal tie is the default connection, e.g.~Facebook. 
Failing to recognise Twitter's eccentric topological structure would make the approach less generalisable. 
Focusing on Twitter leads to a more encompassing framework that can be mapped to other networks. 

\begin{figure}[!b]
    \centering
    \includegraphics[width=0.9\textwidth]{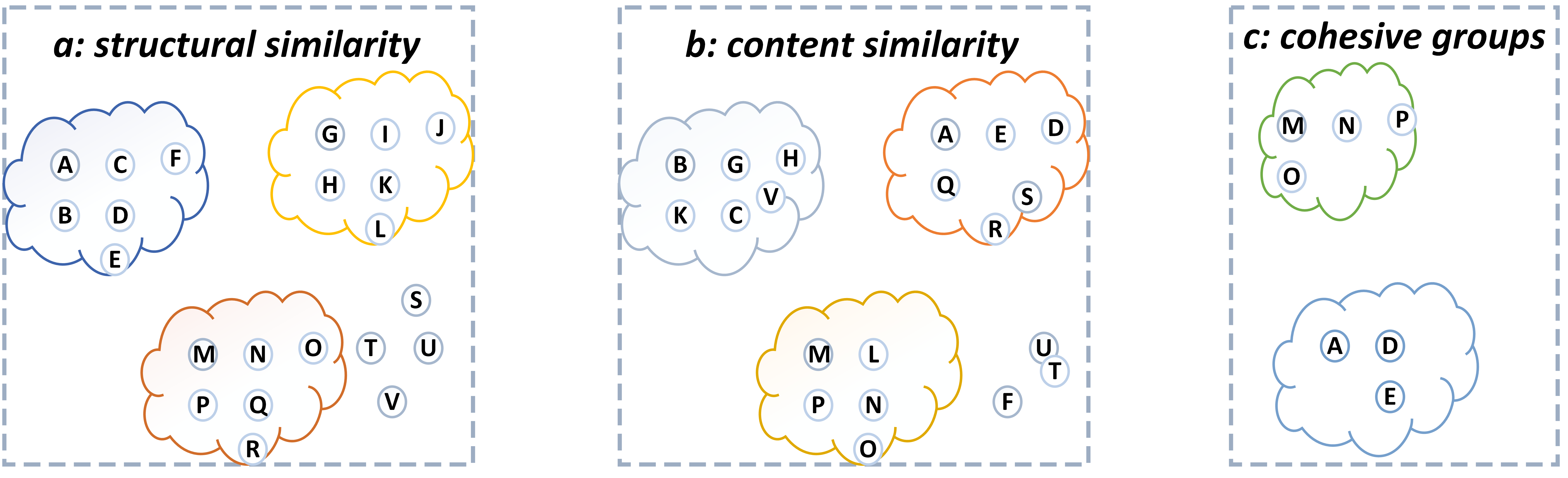}
    \caption[Clustering of users based on factors]{Node clustering in MCT in three stages, according to  
    (a) structural similarity, 
    (b) \textit{content} or \textit{textual} aspect similarity, and  
    (c) joint \textit{structural} and \textit{content} similarity.}
    \label{fig:strutural-content-clusters}
\end{figure}

MCT measures similarity within a community of users using local and global information, modelling \textit{structural} and intrinsic \textit{textual} features jointly. 
In Figure~\ref{fig:strutural-content-clusters}(a) and (b) user communities exist based on \textit{structural} and  \textit{content} or \textit{textual} similarity, respectively. 
Users under the structural component, a form of a \textit{state-type tie}, are related based on reciprocal ties, which are rare in Twitter, and the community is more cohesive than the community of users based on content or textual similarity, a form of an \textit{event-type tie}. 
A more cohesive community is the one that recognises both structural and content similarity, in Figure~\ref{fig:strutural-content-clusters}(c). 
Intuitively, the degree of cohesiveness varies across different communities: 
a community based on both modalities is the most cohesive, followed by a community based on high structural similarity but low or no content or textual similarity. 
Finally, the least cohesive community exhibits high similarity in the textual component but low or no structural similarity. 
Groups of structurally similar nodes are analysed by \textit{spectral clustering}, which involves a series of methods ranging from adjacency and affinity matrices to dimensionality reduction. 
The textual component complements the structural aspect through a form of document-pivot clustering, in which weights are assigned to features in the document according to a weighing scheme \cite{allan1998line,yang2001study,brants2003system,fung2005parameter}. 

\subsection{Contributions}
\label{sec:contributions}

MCT relies on reciprocal ties, based on the assumption that combining structural and textual features offers a more cohesive community representation. 
Our contributions are two-fold: 

\paragraph{A new dimension to the detection of cohesive communities}
The ability to follow anyone on Twitter results in many unidirectional connections between socially unrelated users, affecting clustering and the integrity of online content. 
% \item \textit{addressing challenges in identifying large scale reciprocal ties:} 
To counter the challenging and time-consuming task of collecting large scale reciprocal ties on Twitter, we proposed a strategy that returns the likelihood of reciprocity among users. 
As a result, the detection of socially cohesive communities is enhanced, providing a useful analysis tool and strengthening the validity of online content. 
Moreover, by identifying communities of users with a strong cohesion, a well-informed recommendation that recognises structural and textual similarities can be achieved.

\begin{table}[t]
    \centering
    \footnotesize
    \caption[Summary of notations and descriptions]{Relevant notations and their corresponding descriptions}
    \label{tab:notations-and-description}
    \begin{tabular}{l l}    \hline
        \textbf{Notation} & \textbf{Description}\\  \hline
        $\mathcal{D}$ & network data \\
        $\mathcal{V}$ and $\mathcal{E}$ & sets of nodes and edges, respectively\\
        $m_{v_i}$ &  denotes the network of a user $v_i$\\
        $fr_{v_i}$ and $fl_{v_i}$ &  denote sets of friends and followers of user $v_i$\\
        $\kappa \in m_{v_i} $ & set of reciprocal ties of user $v_i$ \\
        $\mathcal{A}_f$ & set of all possible attributes of a node\\
        $\mathcal{X}_{f}$ & set of features inducing reciprocity\\
        %$\epsilon_{ui}$ & error term in the proposed model $M$ \\
        $a \succ b$ & a binary relation between $a$ and $b$ \\
        $\exp{x}$ or $e^x$ & base of a natural logarithm, $ln$, $\exp{(lnx)}=x$; $e \approx 2.71828$ \\
        $d_i, p_i, q_i, u_i$ & respective $ith$ component of vectors, $\mathbf{D, P,Q,U}$ \\
        %$\phi$ & similarity function \\
        $\tau$ & a predefined threshold for comparison, e.g. $\tau \geq 0.5$ \\
        $\mathcal{S}_r$ and $\mathcal{T}_r$ & sets of structurally-related and textually-related nodes, respectively \\    \hline
    \end{tabular}
\end{table}

\paragraph{A bi-modal community detection approach} 
MCT addresses the problem of structurally unrelated users by adding a layer of social cohesion to existing community detection methods. 
Specifically, MCT advances existing techniques through: 
(a) an in-depth utilisation of a bi-modal source of information for community detection, 
(b) detection of network communities at various levels, 
(c) a robust and scalable community detection algorithm that is less prone to noise in the network data, and 
(d) an intuitive interpretation of the detected communities. 

\vspace{3mm}
The remaining of this paper is structured as follows. 
Section~\ref{sec:background} provides background details and related work. Section~\ref{sec:the-mct-framework} formulates the problem and describes the MCT framework. 
We describe the experimentation process in Section~\ref{sec:experimentation} and discuss the main observations in Section~\ref{sec:discussion}. 
Finally, Section~\ref{sec:conclusion} concludes the study and provides some considerations for future work. 
For ease of referencing, Table~\ref{tab:notations-and-description} summarises the notations used. 

%{\color{green!65!black} GREENISH TEXT}

%% file: p2-background.tex
\section{Background} 
\label{sec:background}
Humans can effortlessly abstract complex phenomena, but efficiently automating the process is daunting partly due to the multidimensional nature of clustering data \cite{bishop2006pattern}. In this section, we review relevant topics and studies associated with clustering and community detection tasks. 

\subsection{Network and community structure} 

A network comprises of heterogeneous nodes connected via edges. 
The topological structure of networks and other quantities related to them are useful in understanding complex networks across numerous domains \cite{albert2002statistical}. 
Various levels of relationship forms in networks have been analysed, from the structure of microscopic organisms to complex networks, such as the internet \cite{scott1988social,watts1998collective,albert2002statistical}. 
Complex networks were once considered to as random and the classic random graph model \cite{erdos1960evolution} was the standard analysis tool until regular patterns in various networks were discovered, e.g., via statistical analysis.
Fundamentally, network complexity \cite{watts2007influentials} is defined by: 
(a) \textit{Clustering coefficient} quantifies the probability of a node to be clustered, assuming that users with common friends are likely to know each other.
(b) \textit{Degree distribution} quantifies the probabilistic distribution of nodes.  
(c) \textit{Small-worldliness} is a network property associated with short path-length, i.e., many structured short-range connections and few random long-range ones, and network diameter that is exponentially less than its size \citep{watts1998collective}. 
%Social Networks

In \textit{social media}, communication happen at various layers of granularity and intensity: global or local, positive or negative, influential or not. 
In contrast with the early unidirectional \textit{two-step} communication model, where few users serve as intermediaries between mass communication and the public \cite{katz2017personal}, the design of social media allows users to generate and consume information. 
On social media, communication follows the influence network model, enabling multi-way flow, where users can simultaneously generate and consume information \cite{watts2007influentials}. 
Twitter is dominated by influential users, logically dividing a clique of content pushers and consumers, resembling the two-step flow model \cite{katz2017personal}. 
This division is strengthened by strategies, such as content promotion, that entice users to engage more, and to follow or add friends.
Using these strategies social media users can increase their network of friends, generating more value to the platforms. 
A social media network is the synthesis of many user communities, and identifying these structures is a vital task. 
Because members of a community are highly similar to each other and less or not at all similar to members of other communities \cite{sundaram2012understanding,lancichinetti2009detecting,newman2004detecting}, a \textit{community structure} has densely connected node groups and sparser connections to other communities \cite{newman2004fast}. 
Thus, community identification involves prediction and quantification tasks to detect the relevant structures and their characteristics \cite{chen2009detecting}. 
Selecting an effective similarity measure is crucial as it is allows a clustering algorithm to identify groups and affects the \textit{signal-to-noise-ratio} within the instance matrix \cite{lawson2012population}. 

\subsection{Related work}
\label{sec:related-work}
Network partitioning has attracted interest from various domains of expertise, hence diverse strategies have been put forward to identify relevant communities embedded in a network. 

\paragraph{Clustering and Community Detection} 

Often clustering and community detection are used interchangeably in the literature. 
Clustering mostly focuses on a single modality, e.g., using node attributes to group network objects, whereas community detection focuses on network structure as a function of connectivity involving social interaction. 
As a form of dimensionality reduction, clustering entails unsupervised partitioning a network into groups of related objects using a domain-specific scoring function and maximising in-group similarity. % and a community detection involves prediction and quantification to identify network structures. Clustering tasks typically aim to discover meaningful inherent patterns without prior knowledge of the structure. 
There are two principal lines of research in this direction: graph partitioning and hierarchical modelling \cite{newman2004detecting}. 
We follow this classification, as it reflects the approach in this study.  
Methods can also be divided in dimensionality reduction based ones, and 
graph partitioning (hierarchical or not \cite{manning1999foundations}) and hierarchical ones \cite{aggarwal2014evolutionary}. 

\subsubsection{Graph-based and hierarchical methods}

Graph-based clustering assumes that a community structure exists in the network and attempts to discover it using specific techniques.
Graph partitioning divides the network into predefined node groups and suits applications where the number and size of groups are known, e.g., in parallelisation of computing processors. 
The approach may involve hierarchical agglomerative clustering \cite{pons2006computing} following a random walk model \cite{erdos1960evolution}, or based on modularity \cite{newman2004fast} optimisation, such as in the Louvain detection algorithm \cite{blondel2008fast}. 
The clustering method can be based on \textit{iterative bisection}, which divides the network optimally into two parts and repeats until the required number of partitions is reached \cite{newman2006modularity}. 
The modularity measure measures community strength and detects groups, assuming that community structures correspond to an interesting statistical arrangement of edges.
Positive values indicate the presence of community structures, i.e., that nodes within a community are more tightly connected than by chance \cite{newman2006modularity}. 
The modularity value of real networks ranges from $.3$ to $.7$. 
The higher the score, the more cohesive the community structure \cite{newman2004detecting}. 
Predefining the maximum bisection size is required, which may affect performance. 
Metrics such as \textit{betweenness} or \textit{shortest loop edges}, are central to the operation of algorithms that process graphs to detect groups of similar nodes \cite{pothen1990partitioning}. 

Hierarchical modelling follows a different technical approach from graph-based clustering. 
Assuming that there are natural subgroups in a network, hierarchical clustering utilises a similarity measure, such as Euclidean distance or Pearson correlation to analyse the network \cite{scott1988social}. 
In particular, pairwise node similarities are computed and nodes are iteratively and deterministically assigned to clusters. 
Commonly, similar clusters are iteratively merged into larger ones \cite{aggarwal2014evolutionary}. 
%\paragraph{Model-based clustering} 
Furthermore, categorisation based on a generative or model-based and discriminative or similarity-based is used in the literature. 
Model-based or generative clustering algorithms, e.g., Latent Dirichlet Allocation (LDA) \citep{blei2003latent,balasubramanyan2011block,yan2013biterm}, are a form of Expectation Maximisation (EM) that aim to learn a generative model from data segments, where each model represents a cluster \citep{berkhin2006survey}. 
The EM-based models estimate the maximum likelihood of data-points to belong to a cluster and is suitable for incomplete data. 
On the other hand, similarity-based clustering algorithms are based on optimising a scoring function that is used to compute pairwise similarity between data-points. This form of clustering follows hierarchical agglomerative clustering or block modelling. 

\subsubsection{Multiview and bi-modal clustering } 
Multiview and bi-modal techniques aim to improve clustering performance using multiple independent data sources; thus, multiview clustering relies on data that can be split into independent sub-features or attributes \cite{chaudhuri2009multi,liu2013multi}. %It is applicable where node attributes can be divided into sub-groups, and each can be used for independent learning. 
For instance, a \textit{web page} can be described by its textual content and pages that link to it \cite{bickel2004multi,chao2017survey}. 
The advantages of multiview clustering over its single view counterpart has been investigated using algorithms based on \textit{K-means} and \textit{Expectation Maximisation} \citep{bickel2004multi}. 

Bi-modal clustering technique is based on the fact that network communities are formed around two primary modalities or information sources: \textit{network structure} and node \textit{attributes}. 
In many cases, structural and textual aspects evolve simultaneously and communities are discovered according to the nodes' similarity degree vis-\'{a}-vis those two aspects. 
Until recently, studies mostly focused on one aspect, not both \cite{yang2013community,balasubramanyan2011block,leskovec2012learning,ester2006joint,zhou2009graph}. 
A study closely related to our approach, proposes a generative model for networks with node attributes \cite{yang2013community}. 
However, the depth of the features, especially the nodes' attributes, is shallow and the node attribute (\textit{hashtag}) is insufficient to analyse the depth of similarity between network entities in a complex environment such as Twitter in which the structural component is not fully captured due to reliance on directed edges. % hence lacking/devoid of the necessary attributes to allow for an in-depth structural insight.  
The connected k-centre approach employs both structural and attribute information for a given network partition \citep{ester2006joint}. 
The problem is NP-hard, leading to many heuristics. 
Similarly, SA-cluster method combines structural and attributes' similarities for community detection by partitioning a network into cohesive k-clusters with structural and attribute information using a distance metric to estimate pairwise node similarity or closeness \citep{zhou2009graph}. %Then, a random walk model identifies relevant nodes in each node's neighbourhood for clustering. 

%Apart from the bi-modal approach, community detection in networks with edge uncertainty or incomplete information is getting traction. 
Conventional methods, such as normalised cut \cite{shi2000normalized} and modularity \cite{newman2006modularity}, are based on topological structures. 
However, many networks come with incomplete information, e.g., a \textit{terrorist network} or food web \cite{lin2012community}; thus, community detection in networks with edge uncertainty or incomplete information is getting traction. 
Inferring links in incomplete networks is challenging, because the information is usually localised within a small, linked group. 
The full wealth of data has been used to learn a generalisable distance metric to complete the missing information \citep{lin2012community}. 
However, this approach is too complicated and does not account for the breadth required in \textit{textual} aspects in networks with many transient connections, such as Twitter\footnote{As shown in Figure~\ref{fig:Twitter-ecosystem}, Twitter communities are formed based on many factors.}. 
The MCT is a two-stage clustering technique that recognises different modalities as information sources; it incorporates multiview aspects at various levels, structural and textual, using independent features. 
%The MCT is similar to multiview clustering, in which multiple features are utilised to improve clustering \cite{chaudhuri2009multi,liu2013multi}.
%#############################################

%% file: p3-microcosm-detection.tex
\section{MCT framework} 
\label{sec:the-mct-framework} 

\begin{figure}[t]
    \centering
    \includegraphics[width=0.5\textwidth]{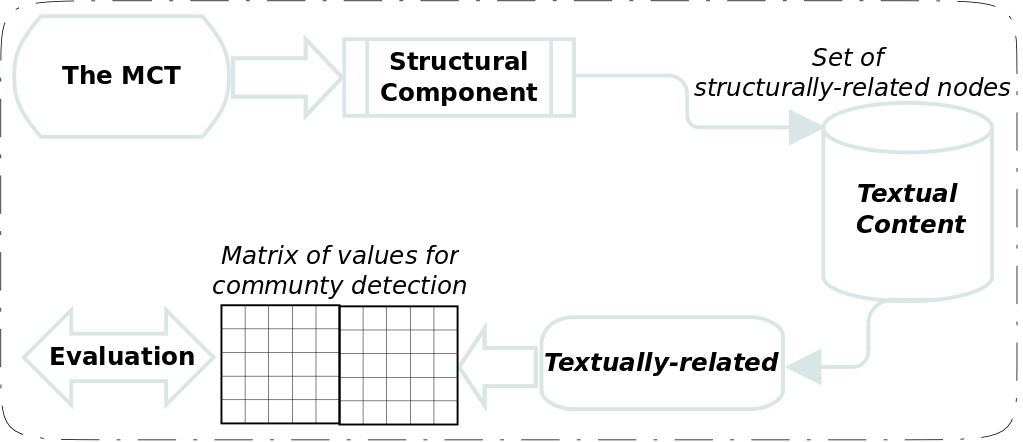}
    \caption[MCT execution pipeline]{
    Execution pipeline of the MCT framework - 
    The structural component processes a collection of structurally similar nodes, promoting group formation among them. 
    Then, \textit{textual analysis} of structurally related node content identifies groups according to discussion topics. 
    MCT combines \textit{state-type} and \textit{event-type} ties.} 
    \label{fig:mct-pipeline}
\end{figure}

Noting that nodes in a community are highly similar and edges among communities are infrequent, community detection is usually formalised to identify network partitions that satisfy specific requirements. 
The problem focuses on detecting smaller groups with high similarities, using a joint similarity function that considers global and local information as a two-stage process comprising of structural and textual components (shown in Figure~\ref{fig:strutural-content-clusters}). 
Figure~\ref{fig:mct-pipeline} shows a block diagram of the stages in the MCT framework. 

\subsection{Structural component}
\label{sec:structural-component} %groundtruth and predicted data ... 

\begin{figure}[!b]
    \centering
    \includegraphics[width=0.65\linewidth]{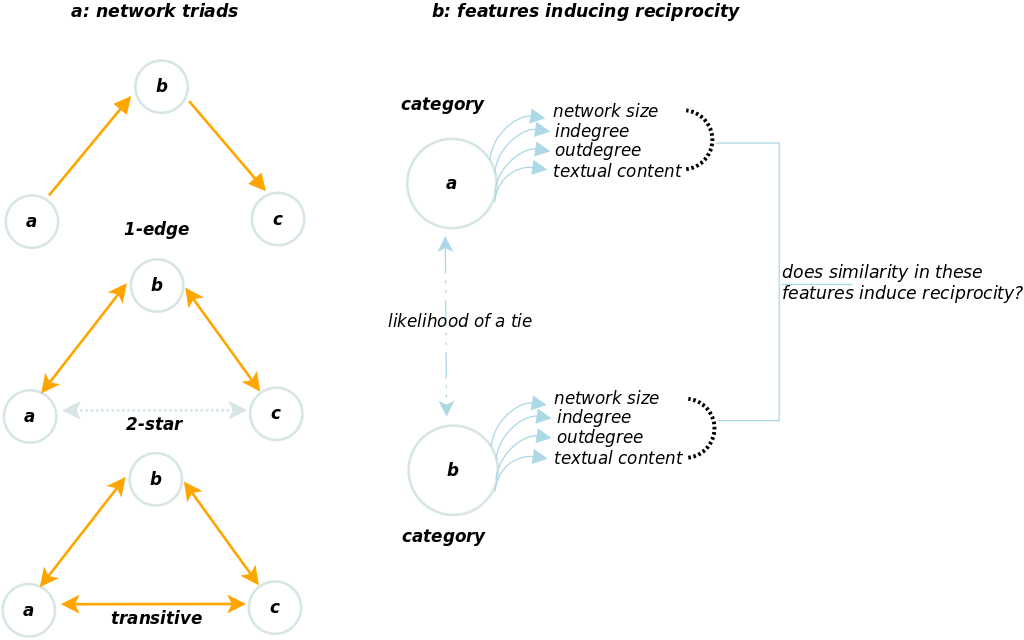}
    \vspace{-3mm}
    \caption[Triads in a network]{(a) Possible social ties in a network triad - Each node is associated to a set of nodes with a directed or reciprocal tie. 
    % In \textit{sub-figure (b)}, $a$ and $b$ denote nodes and their corresponding network compositions given by $N_a$ and $N_b$, e.g.~${a_1,a_2,a_3, ..., a_n} \in N_a$. Thus, a reciprocal tie is established if $\exists a_i \in N_a: a \in N_{a_i}$, where $N_{a_i}$ denotes the network composition of node $a_i$. 
    (b) An example dyad and the features responsible for tie formation between users on Twitter. A probability score is assigned to each feature, to discover their inter-dependencies and enable reciprocal ties.}
    \label{fig:tie-formation-and-strength}
\end{figure}

The structural component is based on dyadic, pairwise edge between two users, and transitive ties, which are the basic forms of establishing reciprocal ties in social networks. We aim to identify groups of users with true reciprocal relationships at dyadic and transitive level. 
Transitivity expresses the social preference to be friends with a \textit{friend-of-a-friend} and has been characterised as a peculiar network feature \cite{watts1998collective}. 
Transitive ties are synonymous to \textit{Simmelian ties}, strong social relationships among three or more individuals, which are vital in understanding a network's social tie structure \cite{granovetter1977strength}. 
Our approach assumes that community detection or clustering methods that take reciprocal ties into account offer a more cohesive community representation. %Our analysis of Twitter datasets concluded that true reciprocal ties are rare. 
Our analysis of Twitter datasets concluded that true reciprocal ties are rare. Thus, we use a method that strengthens the possibility of finding Twitter users with reciprocal ties. A user with many reciprocated ties can represent a \textit{microcosms}, allowing to analyse a user group as a unit. 
Research in social science suggest that users compare themselves with one another and adopt similar behaviour with users similar to them \cite{brass1998relationships}. 
Homophily on Twitter can be interpreted as a reciprocal relationship among users. 
Noting this insight and the inspiration drawn from social homophily, we argue that users with similar profiles are more likely to connect on Twitter. Therefore, structural equivalence is mapped to a \textit{state-type tie} to infer structural similarity according to the node's attributes. 
Figure~\ref{fig:tie-formation-and-strength} shows features that contribute in finding structurally related nodes.

\subsubsection{Modelling structural clusters}
\label{sec:structurally-related-clusters}

\paragraph{Definitions} 
This section begins with the definitions of relevant concepts and terms in the implementation. Table~\ref{tab:notations-and-description} provides a summary of all relevant notations used in the study. %that are central to the structural aspect.
\begin{itemize}
    \item[-] \emph{Network data $\mathcal{D}$} consists of sets of nodes, ${v_1,v_2,...,v_m} \in \mathcal{V}$, and edges, ${e_1,e_2,...,e_n} \in \mathcal{E}$. 
    Each node is described by its structural and textual features, as shown in Figure~\ref{fig:tie-formation-and-strength}.
    \item[-] \emph{Dyadic and transitive ties:} a relation, $\succ$, between two nodes $v_i,v_j \in \mathcal{D}$ is dyadic\footnote{Dyadic tie, pairwise, 2-star or binary relations are used interchangeable in this study.} if $v_i$ follows $v_j$ and vice versa, i.e.~$v_i \succ v_j \Leftrightarrow \forall v_i, v_j \in \mathcal{D}$. 
    In this context, $v_i$ follows $v_j$ is a directed relationship; if $v_j$ follows $v_i$ back, it is undirected; see some examples in Figure~\ref{fig2:ab-connection}. 
    \begin{figure}[b]
        \centering
        \includegraphics[scale=0.72]{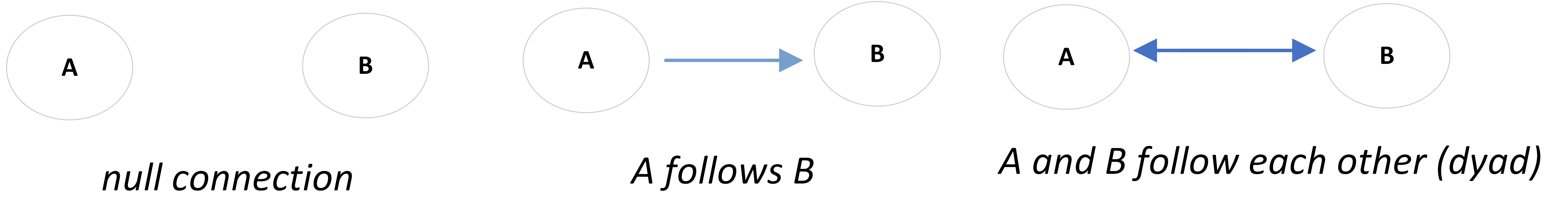}
        \vspace{-4mm}
        \caption{Examples of possible relationships between pairs}
        \label{fig2:ab-connection}
    \end{figure}
    A transitive or \textit{Simmelian tie} is a social relationship within groups of three or more. 
    A binary relation, $\succ$, over a set $\mathcal{D}$ is \textit{transitive}: 
    $v_i\succ v_j \mbox{, } v_j\succ v_k \Leftrightarrow v_i \succ v_k \mbox{, } \forall v_i, v_j,v_k \in \mathcal{D}$. 
    \item[-] \emph{Prediction features and reciprocity:} To identify structurally related nodes, we use features, such as \textit{indegree (ind)}, the number of incoming edges to a node; \textit{outdegree (out)}, the number of outgoing edges from a node; and \textit{category (cat)}, indicating if a node is \textit{verified} or \textit{unverified}. 
    Account verification ascertains legitimacy of the account holder. 
    $\mathcal{A}_f$ denotes the set of all possible node features, which can be used to extract feature subsets, $\mathcal{X}_f = \{ind, out, cat\} \subset \mathcal{A}_f$, from a profile, as in Figure~\ref{fig:tie-formation-and-strength}. 
    % \footnote{The figure depicts easy-to-access features that allow users to quickly decide about a follow request.} 
    The features of a node pair, $v_i,v_j$, are denoted by: $\mathcal{X}_{f_{v_i}} = \{ind_{v_i}, out_{v_i}, cat_{v_i}\}$ and $\mathcal{X}_{f_{v_j}}= \{ind_{v_j}, out_{v_j}, cat_{v_j}\}$ and are used for training models that predict the likelihood of \textit{reciprocity}. 
    %\item[-] \emph{Structural similarity:} Any pairs of nodes $v_i$ and $v_j$ are structurally-similar or -related $\mathcal{S}_r$, if their degree of reciprocity, given by $p(R_{v_i,v_j})$, is greater than a predefined threshold $\tau$. It follows that $\mathcal{S}_r: \forall v_i \in \mathcal{S}_r ~\exists v_j \mbox{ such that } p(R_{v_i,v_j}) \geq \tau \hspace{1mm} \mbox{ where } \mathcal{S}_r\subset \mathcal{D}$. 
\end{itemize}

\paragraph{Nodes reciprocity and constant error}

%To identify structurally-related nodes, we begin with a high-level grouping of nodes. %into \textit{network-communities} and \textit{reciprocal-communities} according to \textit{network size} and \textit{reciprocity} respectively. 
We predict node sets with dyadic ties on Twitter using the approach in \cite{inuwa2019analysis}, and we use them for clustering. 
The formulation assumes that reciprocity is based on the features that can induce friendship, in Figure~\ref{fig:tie-formation-and-strength}(b). 
Nodes reciprocity hypothesises that the decision to reciprocate or establish friendship correlates with the idea of homophily and structural equivalence. 
Reciprocal ties are predicted based on these concepts between node pairs. 
%As held by \citet{ahn2010link}, where attributes' similarities have been used for community detection tasks, the probability of reciprocity resulting from matching node features is based on high similarity of attributes.  

Consider the sets of nodes, $\mathcal{V}$, and edges, $\mathcal{E}$.
The likelihood of \textit{reciprocity}, $p(R_{v_i,v_j}) \mbox{, } \forall v_i,v_j \in \mathcal{V}$) involved in the computation of reciprocal units (see Section~\ref{sec:reciprocal-units}) is described by Eq.~\ref{eq:jaccard-index} to \ref{eq:probable-reciprocity}, leading to \textit{reciprocal-communities} $\mathcal{C}_{rc}$. 
The features of a node pair, $v_i,v_j$, for comparison are denoted by: $\mathcal{X}_{f_{v_i}}$ and $\mathcal{X}_{f_{v_j}}$, such that the ratio of the attributes, e.g. \textit{ind} or \textit{out}, is a real value quantity given by: $\frac{ind_{v_i}}{ind_{v_j}} \in \mathbb{R} \hspace{3mm} \forall f_r \in \mathcal{X}_{f_{v_i,v_j}}$. %$$\frac{ind_{v_i}}{ind_{v_j}} \in \mathbb{R} \hspace{10mm} \forall f_r \in \mathcal{X}_{f_{v_i,v_j}}$$ 
If the comparison evaluates to a value in $[0.75, 1.25]$, the pairs are assumed to have similar attributes (1), otherwise dissimilar attributes (0). 
The interval is to allow extra freedom for minor discrepancies between the features. 
For instance, if the ratio equals $1.0$, the pair has identical attributes, which is useful in analysing aspects of homophily and social equivalence. 
The binary feature comparison values are used to compute the overall similarity using the \textit{Jaccard Index, J}:
    \begin{equation}
        \label{eq:jaccard-index}
        J(\mathcal{X}_{f_{v_i}},\mathcal{X}_{f_{v_j}}) = \frac{|\mathcal{X}_{f_{v_i}} \cap \mathcal{X}_{f_{v_j}}|}{|\mathcal{X}_{f_{v_i}} \cup \mathcal{X}_{f_{v_j}}|}
    \end{equation}
Moreover, modelling reciprocity or the response to a \textit{friendship} request is associated with a decision error, that can be quantified based on \textit{response probability}. %or \textit{probabilistic preference model} 
Response probability applied to cases where a feature set for a decision involves a constant probability of error (Eq.~\ref{eq:error-term}) in the choice \cite{marley2016choice}. 
Thus, the probability of \textit{reciprocity} between pairs based on the similarities in their features $J(v_i,v_j)$ (Eq.~\ref{eq:jaccard-index}) is given by \cite{inuwa2019simmelian}: %The error $\epsilon_{v_i,v_j}$ (Eq.~\ref{eq:error-term}), associated with each prediction is expressed as: 
    \begin{equation}
        \label{eq:error-term}
        \epsilon_{v_i,v_j} = \frac{1}{\zeta \times (1+ \log(J(v_i,v_j)+\zeta))}
    \end{equation}
The \textit{constant error term}, $\zeta$, is assigned the value of $1/3$ and the final relation is given by:
    \begin{equation}
        \label{eq:probable-reciprocity}
        p(R_{v_i,v_j}) = \frac{1}{1+ \exp{^\varphi}}
    \end{equation}
where $\varphi = -\log(\epsilon_{v_i,v_j}+J(v_i,v_j))\times(\epsilon_{v_i,v_j}+J(v_i,v_j))$. 
Any node pair $v_i$, $v_j$ are structurally similar or related $\mathcal{S}_r$, if their degree of reciprocity (Eq.~\ref{eq:probable-reciprocity}) is greater than a predefined threshold $\tau$. It follows that $\mathcal{S}_r: \forall v_i \in \mathcal{S}_r ~\exists v_j \mbox{, such that } p(R_{v_i,v_j}) \geq \tau \hspace{1mm} \mbox{, where } \mathcal{S}_r\subset \mathcal{D}$. %The formal process is given by \textit{Algorithm f-sim} (Algorithm~\ref{alg:f-sim}). 

\paragraph{Collection of structurally related nodes}

Eq.~\ref{eq:probable-reciprocity} allows to identify as many node sets with a high likelihood of establishing reciprocal ties, thus adding a layer of social cohesion to the MCT framework. 
Identifying groups of structurally related nodes begins with a high-level aggregation of nodes according to network size (for \textit{network-communities}) and reciprocity (for \textit{reciprocal-communities}). 
This led to the question \textit{what does it mean for nodes to be structurally related?} 
As a simplistic example, consider a finite set $\mathcal{V}_{13}$ that contains 13 nodes: $\{v_1, \dotsi, v_{13}\} \in \mathcal{V}_{13}$. 
After executing \textit{Algorithm f-sim} (Algorithm~\ref{alg:f-sim}), which predicts the likelihood of reciprocity, the following pairs of nodes are structurally-similar or related\footnote{The symbol '$\sim$' is used to denote structural similarity between pairs.}: $v_1 \sim v_2, v_1 \sim v_3, v_1 \sim v_5, v_2 \sim v_4, v_2 \sim v_5, v_2 \sim v_9, v_3 \sim v_{11}$. 
Accordingly, three structurally related communities can be identified: 
$c_1 = \{v_1,v_2,v_3,v_5\}$, $c_2 = \{v_2,v_4,v_5,v_9\}$ and $c_3 = \{v_3,v_9\}$. 
The communities can be expressed in a matrix form for spectral analysis. 
Matrix entries can be based on states, such as the \textit{reciprocity potential} of nodes defined as the ratio of \textit{outdegree} over \textit{network size}. 
 %Algorithm f-sim
    \begin{algorithm}[t]
    \small
    \caption{\emph{: Algorithm f-sim} returns the likelihood of reciprocity between pairs.}
    \label{alg:f-sim}
        \begin{algorithmic}[1]
        \State \textbf{Initialisation:} $\{\} \longleftarrow \mathcal{S}_r; \{\} \longleftarrow \mathcal{S}_u$
        \State \textbf{Input:} a finite collection of network data $\mathcal{D}$
        \While {$\mathcal{D} \neq \emptyset$}
        \State $\forall v_i, v_j \in \mathcal{D}$, compute $p(R_{v_i,v_j})$ using Eq.~\ref{eq:probable-reciprocity} $\hspace{12.5mm} \triangleright~ v_i\neq v_j$
        \If{$p(R_{v_i,v_j}) \geq \tau$} $\hspace{46.5mm} \triangleright ~\tau$, \textit{a predefined threshold}
        \State $\mathcal{S}_r \gets (v_i,v_j) \hspace{54mm} \triangleright$ \emph{structurally related}
        \Else
        \State $\mathcal{S}_u\gets (v_i,v_j) \hspace{54mm} \triangleright$ \emph{structurally unrelated}
        \EndIf
        \EndWhile
        \State \textbf{Output:}
        \State $\hspace{5mm} \mathcal{S}_r,\mathcal{S}_u, \mathcal{M}_{A_{i,j}}^{n\times n}\hspace{59mm}\triangleright~$\emph{adjacency matrix} $\mathcal{M}_{A_{i,j}}^{n \times n}$
        \end{algorithmic}
    \normalsize
    \end{algorithm}

\paragraph{\textbf{Spectral analysis}}  

Since structurally related nodes can be easily transformed into a graph, we apply \textit{spectral analysis} to identify clusters and enabling \textit{Sociometry}\footnote{Metrics such as structural hole, homophily, and centrality metrics -- degree, closeness, betweenness}, a means to measure social relationships \cite{wasserman1994social}. %can be applied on the detected clusters. 
Spectral clustering involves operations ranging from the construction of adjacency or affinity matrices to clustering in a reduced dimension \cite{han2011data}. 
We construct the \textit{adjacency}, \textit{similarity} and \textit{degree} matrices based on ground-truth data and Eq.~\ref{eq:probable-reciprocity}. 
The adjacency matrix $\mathcal{M}_{A_{i,j}}$ (Eq.~\ref{eq:adjacency-matrix}) encodes the structural similarities among node pairs. 
%(represented by the \textit{row} and \textit{column} indices) 
Presence or absence of reciprocity is marked with 1 or 0, respectively. 
    \begin{equation}
        \label{eq:adjacency-matrix}
        \mathcal{M}_{A_{i,j}} =
        \begin{cases}
            1     & \quad \text{if } p(R_{v_i,v_j}) \geq \tau\\
            0     & \quad \text{if } otherwise
        \end{cases}
    \end{equation}
The \textit{degree matrix}, $\mathcal{M}_D$, is a diagonal matrix obtained by summing the entries in Eq.~\ref{eq:adjacency-matrix} across the rows, in which entry $i,i$ denotes the degree or number of edges to node $i$. 
Thus, each entry in the diagonal $d_i$ (Eq.~\ref{eq:degree-matrix}), of matrix $\mathcal{M}_D$ is defined by:
            \begin{equation}
                \label{eq:degree-matrix}
                d_i = \sum_{\{j|(i,j) \in E\}} p(R_{v_i,v_j}) \geq \tau
            \end{equation}
Subtracting the adjacency matrix, $\mathcal{M}_{A_{i,j}}$, from the degree matrix, $\mathcal{M}_D$, gives the graph \textit{Laplacian}, $\mathcal{M}_L = \mathcal{M}_D - \mathcal{M}_A$, whose \textit{eigenvectors} and \textit{eigenvalues} offer informative features for clustering. 
Diagonal entries in Eq.~\ref{eq:graph-laplacian} denote the degree of nodes, off-diagonal negative entries ($-p(R_{v_i,v_j})$) represent edges between node pairs and zero entries signify no edges. 
\begin{equation}
    \label{eq:graph-laplacian}
    \mathcal{M}_{L_{i,j}} =
        \begin{cases}
            d_i       & \quad \text{if } i=j \\
            -p(R_{v_i,v_j}) & \quad \text{if } p(R_{v_i,v_j}) \geq \tau \hspace{10mm} \triangleright edge(i,j)~exists \\
            0     & \quad \text{if } otherwise
        \end{cases}
\end{equation}

\subsubsection{Structural optimisation}%Optimisation of structural clusters}   

Given a set of structurally related nodes, hidden local communities can be uncovered via matrix decomposition on the following matrices of interactions and corresponding dimensions: 
\vspace{-\topsep}
\begin{itemize}
  \setlength{\parskip}{0pt}
  \setlength{\itemsep}{0pt plus 1pt}
    \item[-] $\mathcal{M}_{c_{ns}} \mapsto n \times p$: a matrix of nodes according to a concept, e.g., network size 
    \item[-] $\mathcal{M}_{c_{vr}} \mapsto n \times k$: a matrix of nodes according to reciprocity 
    \item[-] $\mathcal{M}_{c_{nr}} \mapsto p \times k$: a matrix of high-level and local communities 
\end{itemize}
\vspace{-\topsep}
The network-communities matrix, $\mathcal{M}_{C_{ns}}^{n\times p}$, is decomposed into its approximate constituents:
%$\mathcal{M}_{C_{ns}} \approx \mathcal{M}_{vr}\mathcal{M}_{C_{nr}}^T$. 
%(Eq.~\ref{eq:structural-matrix}): 
    \begin{equation}
       \label{eq:structural-matrix}
       \mathcal{M}_{C_{ns}} \approx \mathcal{M}_{vr}\mathcal{M}_{C_{nr}}^T
   \end{equation}
Interpretability is desirable in the MCT framework, hence, the decomposition follows a non-negative matrix factorisation (NMF) scheme \cite{lee1999learning}. 
NMF provides an intuitive factorisation, in which non-negative constraints are imposed on the optimisation parameters \cite{aggarwal2018machine}.  
    \begin{equation}
        \label{eq:structural-optimisation}
            min._{\mathcal{M}_{c_{vr}},\mathcal{M}_{c_{nr}}}||\mathcal{M}_{c_{ns}} - \mathcal{M}_{c_{vr}}\mathcal{M}_{c_{nr}}^T||_F^2 \mbox{,} \hspace{2mm} \mbox{subject to } \mathcal{M}_{c_{vr}},\mathcal{M}_{c_{nr}} \geq 0
    \end{equation}
We simplify the notation of matrices as follows: $\mathcal{M}_{c_{ns}} \mapsto D$, $\mathcal{M}_{c_{vr}} \mapsto P=[p_{is}]$, $\mathcal{M}_{c_{nr}} \mapsto Q=[q_{js}]$. The formulation in Eq.~\ref{eq:structural-optimisation} allows to consider \textit{Lagrangian relaxation} to optimise the squared Frobenius norm ($||\cdot||_F^2$) of the matrix. %parameters in (Eq.~\ref{eq:structural-optimisation}). 
Consequently, NMF's non-negative constraints are relaxed by introducing the \textit{Lagrangian multipliers}, two new parameters ($\alpha$, $\beta \leq 0$), to the corresponding entries of the optimisation parameters ($P,Q$). 
Accordingly, the objective function $M_{sr}$ is expressed as a \textit{minmax} problem, that requires a simultaneous minimisation over $P, Q$ and maximisation\footnote{This is needed because the Lagrangian multipliers are initialised with negative values.} over all applicable values of $\alpha$ and $\beta$:
    \begin{equation}
        \label{eq:lagrangian-relaxation}
        M_{sr}=||D-PQ^T||_F^2 + \sum_{i=1}^n\sum_{s=1}^k p_{is}\alpha _{is} + \sum_{j=1}^p\sum_{s=1}^k q_{js}\beta _{js}
    \end{equation} 
The optimisation starts by computing the gradient of the relaxed Lagrangian, with respect to the first aspect of the \textit{minmax} (i.e.~minimisation) optimisation variables. 
Although $\alpha$ and $\beta$ offers a degree of flexibility (that comes at a cost), optimal solution requires optimisation conditions to be based on $P$, $Q$ only. 
We apply a handy technique based on Karush-Kuhn-Tucker (KKT) optimality condition \cite{bertsekas1997nonlinear} to eliminate the Lagrangian multipliers. 
The KKT condition suggests that $p_{is}\alpha _{is} = q_{js}\beta _{js}=0$.
Non-negative random values in $(0,1]$ are iteratively assigned to the parameters $P$ and $Q$ based on the following update rule (after simplifying equations as shown in the Appendix): 
    \begin{equation}
        \label{eq:structural-update-rule}
            p_{is}\Leftarrow \frac{(DQ)_{is}p_{is}}{(PQ^TQ)_{is}} \mbox{,} \hspace{2mm} q_{js}\Leftarrow \frac{(D^TP)_{js}q_{js}}{(QP^TP)_{js}}\mbox{,} \hspace{2mm} \forall i \in \{1, \dotsi, n\} \mbox{, } j \in \{1, \dotsi, n\} \mbox{, } s \in \{1, \dotsi, k\}
    \end{equation}
The process of updating $P$ and $Q$ involves comparing their values to the original matrix $D$, aiming in minimising the difference or \textit{error}.
Parameters $p_{is}$ and $q_{js}$ are iteratively updated until convergence. 
A successful iterative update process ensures that the underlying matrices exhibit strong correlations among their respective entries. 

\subsection{Textual component}
\label{sec:textually-related-clusters} 

The textual component applies a form of document-pivot clustering based on weighted features \cite{allan1998line,yang2001study,brants2003system,fung2005parameter}. 
Due to the volume of tweets and their short length, it is difficult to gain a broad perspective about topics. 
Hence, using a single tweet may not provide sufficient information. 
To understand the discussion topics and the degree of similarity among tweets, a fixed number of tweets is extracted from nodes in the structurally related sets, $\mathcal{S}_r$. 
We utilise Latent Dirichlet Allocation (LDA), which has been previously applied for similar tasks \cite{airoldi2008mixed,yan2013biterm,yali2014biterm}. 
LDA is a probabilistic generative model that assigns word distributions to \textit{topics} and topic distributions to \textit{documents} in a corpus, so that the documents represent random mixtures over latent topics \cite{blei2003latent}. 
In this study, tweets collected from each node $v_i \in \mathcal{S}_r$ define a corpus $\mathcal{T}_{v_i}$, whose overall theme is analysed for comparison with other nodes. 

\subsubsection{Modelling textual clusters}
\label{sec:modelling-textual-clusters}
Identifying textually related nodes, $\mathcal{T}_r$, starts by aggregating a finite collection of textual content, $\mathcal{T}$, from each node $v_i$. 
The collection of $k$ tweets produced by node $v_i$ over time, i.e.~$\{t_{i1},t_{i2},t_{i3}, ...,t_{ik}\} \in \mathcal{T}_{v_i}$ consists of a set of $m$ n-gram features $\{f_{i1},f_{i2},f_{i3}, ...., f_{im}\} \in t_i \in \mathcal{T}_{v_i}$.  
Each stream of tweets is preprocessed to extract shingles\footnote{\textit{Shingles} are obtained by removing stopwords and other non-content bearing terms in a tweet.} for transformation following the \textit{term-frequency-inverse document frequency (tf-idf)} weighting scheme \cite{salton1988term}. 
The \textit{tf-idf} vector of a tweet, $\mathbf{v_i}$, can be normalised or not; we apply the $L_2-norm$ given by: $\mathbf{v_i} =\frac{1}{n}\sum_{i=1}^n \frac{\mathbf{v_i}}{||\mathbf{v_i}||_2}$. %\footnote{A normalised vector is a unit vector, i.e., a scaled version of the original, $\mathbf{v}$, by dividing by its length $||\mathbf{v}||$.}.

\paragraph{Similar documents}
The collections of tweets for any node pair are: $\mathcal{T}_{v_i}, \mathcal{T}_{v_j} \in \mathcal{T}_r \subset \mathcal{S}_r$. 
The aggregation scheme ensures that each node has a distinct fingerprint for comparison with others. 
We train the LDA model so that each tweets in the corpus has a finite distribution over topics, and topics have distributions over words. 
The distribution of each tweet, dubbed \textit{anchor tweet}, is used to compare with other tweets to locate the most similar tweets and generate relevant matrices. 
Because LDA-based comparison relies on the probability distributions of tweets, we apply Jensen-Shannon Divergence (JSD), a useful statistical metric, to measure tweet similarity  as the degree of divergence in the respective distributions. 
Unlike Kullback-Leibler Divergence, \textit{JSD} is symmetric, which is crucial in the task of comparing tweets, since similarity should be the same irrespective of the order, i.e.,$X \mapsto Y$ or $Y \mapsto X$ be equal. For example, given two discrete distributions $X$ and $Y$, \textit{JSD} is defined as: 
\begin{equation}
    \label{eq:js-divergence}
    JSD(X||Y) = \frac{1}{2}D(X||\mu) + \frac{1}{2}D(Y||\mu)
\end{equation}
The \textit{JSD} distance measure ($JS_{dist}$) is obtained by squaring its \textit{divergence relation}:
    \begin{equation}
        \label{eq:js-distance}
        JS_{dist} = \sqrt{ \left(JSD(X||Y \right)}
    \end{equation}
%\emph{Textual similarity:} and %\paragraph{Matrix of values}
It follows that any pairs of tweets, $t_i$ and $t_j$, are textually-similar or related, $\mathcal{T}_r$, if their similarity degree, $\phi$, is greater than a predefined threshold $\tau$. 
Thus, $\forall t_i \in \mathcal{T}_r \exists t_j: \phi(t_i,t_j) \geq \tau \mbox{, } \hspace{2mm} \mathcal{T}_r \in \mathcal{S}_r$. 
Because a finite collection of tweets is extracted from each node in $\mathcal{S}_r$, each $t_i \in \mathcal{T}_r$ consists of a node and its set of tweets. 
LDA outputs come in a dense $d \times t$ matrix $\mathcal{M}_{lda}^{d\times t}$, consisting of $d$ tweets and their corresponding $t$ topics. 
Moreover, two matrices apply to the textual component: 
(a) $\mathcal{M}_{vt} \mapsto m \times q$: matrix of $m$ nodes and top $q$ topics, and (b) $\mathcal{M}_{va} \mapsto m \times m:$ affinity matrix of nodes according to topic similarity. 
Consequently, node communities are formed around common discussion topics and the goal to cluster them according to topical similarities, as in $tr(\mathcal{M}_{va}\mathcal{M}_{vt}\mathcal{M}_{vt}^T)$. %Formula \ref{eq:textual-model}. 
Algorithm~\ref{alg:text-sim} describes how to obtain the textually related clusters.
%Using the affinity matrix based on $\mathcal{S}_r$ or $\mathcal{T}_r$, various community detection algorithms can be used to identify relevant partitions. Thus, community detection is based on optimising the joint similarities of $\mathcal{S}_r$ and $\mathcal{T}_r$: $\psi(S_r,T_r) = (\phi(S_r)+\phi(T_r))$. 
% Textual/Content Similarity Algorithm

    \begin{algorithm}[t]
    \small
    \caption{\emph{: Algorithm text-sim} identifies \textit{textually related clusters}}
    \label{alg:text-sim}
        \begin{algorithmic}[1]
        \State \textbf{Initialisation:} $\{\} \longleftarrow \mathcal{T}_r; \{\} \longleftarrow \mathcal{T}_u$
        \State \textbf{Input:} collection of \textit{structurally related nodes} $\mathcal{S}_r$
        \State $\forall v_i \in \mathcal{S}_r, \mbox{get k texts } g(\mathcal{T}_{v_i}) \hspace{40mm} \triangleright g(\mathcal{T}_{v_i})$ \emph{set of k texts of} $v_i$
        \State $\hspace{5mm} \mathbf{x_i} \longleftarrow t_i \in g(\mathcal{T}_{v_i}) \hspace{48mm}\triangleright~$ \emph{get texts vectors }$\mathbf{x_i}$ 
        \State $\hspace{5mm} \mbox{truncate } \mathbf{x_i}\hspace{61mm} \triangleright$ \emph{retain b top terms in vector} $\mathbf{x_i}$
        \State $\hspace{5mm} m(\mathcal{T}_{v_i})=\frac{1}{n}\sum_{i=1}^n\frac{\mathbf{x_i}}{||\mathbf{x_i}||_2} \hspace{39mm}\triangleright~$\emph{mean of $L_2$ normalised} $\mathbf{x_i}$
        \State $\hspace{5mm} LDA(m(\mathcal{T}_{v_i}))\hspace{55mm}\triangleright$ \emph{invoke the LDA on} $m(\mathcal{T}_{v_i})$
        \State $\hspace{5mm} \mathcal{T}_{sim}(\mathcal{T}_{v_i}, \mathcal{T}_{v_j}) = JS_{dist}(\mathcal{T}_{v_i}||\mathcal{T}_{v_j}) \hspace{25mm}\triangleright~$\emph{get similar texts using Eq.~\ref{eq:js-distance}} 
        %\State or, if $m(\mathcal{T}_{v_i})$ is normalised: $$m'=\frac{m(\mathcal{T}_{v_i})}{||m(\mathcal{T}_{v_i})||_2}$$
        \If{$\mathcal{T}_{sim}(\mathcal{T}_{v_i}, \mathcal{T}_{v_j}) \geq \tau$} $\hspace{30mm}$
        \State \emph{update} $\mathcal{T}_r \hspace{62mm} \triangleright$ \emph{textually related}
        \Else
        \State \emph{update} $\mathcal{T}_u \hspace{62mm} \triangleright$ \emph{textually-unrelated}
        \EndIf
        \State \textbf{Output:}
        \State $\hspace{5mm}\mathcal{T}_r, \mathcal{T}_u,\mathcal{M}_{ta}^{m\times m} \hspace{56mm}\triangleright$ \emph{affinity matrix } $\mathcal{M}_{ta}^{m\times m}$
        \end{algorithmic}
    \normalsize
    \end{algorithm}

\subsection{Microcosm detection algorithm}
\label{sec:microcosm-detection-algorithm}

The problem of discovering community structures is modelled as a multilevel clustering task, in which nodes are grouped according to scoring functions. 
Using the affinity matrix based on $\mathcal{S}_r$ or $\mathcal{T}_r$, various algorithms can be used to identify relevant partitions by optimising the separate and joint similarities of $\mathcal{S}_r$ and $\mathcal{T}_r$: $\psi(S_r,T_r) = \phi(S_r)+\phi(T_r)$. 
A cohesive community is a collection of nodes, $\mathcal{V}$, with high degree of similarity structurally and textually. 
Thus, the microcosm detection problem can be formally defined as:
\vspace{-0.5mm}
\begin{quote}
    given a collection of network data $\mathcal{D}$, defined by sets of nodes, $\mathcal{V}$ and edges, $\mathcal{E}$, for each node $v_i \in \mathcal{V}$ consisting of sets of \textit{structural} and \textit{textual} features, the goal is to identify a collection of highly cohesive sub-networks $\mathcal{P}$:
\end{quote}
\vspace{-1.4mm}
    $$\mathcal{P}: \forall v_i \in \mathcal{S}_r ~\exists v_j: p(R_{v_i,v_j}) \geq \tau \mbox{ and }
    \forall t_i \in \mathcal{T}_r\subset \mathcal{S}_r ~\exists t_j: \phi(t_i,t_j) \geq \tau \mbox{, } \hspace{2mm}\mathcal{P} \subset \mathcal{D}$$
% \begin{addmargin}[1em]{2em}
% \end{addmargin}

\noindent The above formulation means that for all node pair in the partition $\mathcal{P}$, both the \textit{structural} and \textit{textual} similarities are greater than their respective threshold $\tau$\footnote{For all experiments, $\tau \geq 0.5$, i.e.~pairs are considered similar (1) if $\tau \geq 0.5$, otherwise dissimilar $(0)$.}. %The goal of a clustering algorithm is to maximise \textit{intra-cluster} and minimise \textit{inter-cluster} similarities among nodes; this ensures a distinctive property for different clusters. Typically, the members of a community are similar in some respect, i.e.~based on a scoring function to cluster nodes. 

\paragraph{Community of related nodes} 
With $\mathcal{S}_f$ and $\mathcal{T}_f$ denoting feature sets of structurally and textually related nodes, $\mathcal{M}_{S_{f}}^{n \times n}$ and $\mathcal{M}_{T_{f}}^{m \times m}$ define an adjacency matrix of the structural component and an affinity matrix based on the textual similarity component, respectively. 
Therefore, for each matrix there exist community sets ($\mathcal{K} \in \mathbb{R}^{n \times k}, \mathcal{Q} \in \mathbb{R}^{m \times q}$), such that $\{k_1, \dotsi, k_n\} \in \mathcal{K}$ denotes possible communities in $\mathcal{M}_{S_{f}}$ and $\{q_1, \dotsi, q_m\} \in \mathcal{Q}$ denotes communities in $\mathcal{M}_{T_{f}}$. 
For a matrix of reciprocal relationships $R\subseteq \mathcal{V \times V}$\footnote{$R$ applies to both $\mathcal{M}_{S_{f}}$ and $\mathcal{M}_{T_{f}}$.} and the associated network data $\mathcal{D}$ ($\mathcal{V},R \in \mathcal{D}$), there exist numerous communities $\{c_1, \dotsi, c_k\} \in \mathcal{C}$ such that $\emptyset \subset c_i\subseteq \mathcal{V}$ and $\mathcal{C}$ denotes a community set. 
With any pair of similar nodes denoted by $v_i \sim v_j \iff \exists c_i \in \mathcal{C}: v_i,v_j \in c_i$, a more socially cohesive node community is formed by identifying overlapping nodes in both $\mathcal{K}$ and $\mathcal{Q}$, through a repetitive partition optimisation. 
Accordingly, the MCT framework contributes into two operational categories: (1) \textit{optimising matrices of values} and (2) \textit{optimising intra-cluster similarity}. 
The MCT can be considered as a multivariate function, made up by structural and textual components, allowing to define an objective function that maximises the overall joint similarity. 
% The \textit{MCT Algorithm}
    \begin{algorithm}[t]
    \small
    \caption{\emph{: Algorithm MCT} identifies local communities known as \textit{microcosms} in a network.}
    \label{alg:mct}
        \begin{algorithmic}[1]
        \State \textbf{Initialisation:} $\{\} \longleftarrow \mathcal{S}_r, \{\} \longleftarrow \mathcal{S}_u, \{\} \longleftarrow \mathcal{T}_r, \{\} \longleftarrow \mathcal{T}_u$
    \State \textbf{Input:} a collection of network data $\mathcal{D}$
    \State structural-component: $\hspace{55mm}\triangleright$ \emph{invoke f-sim (alg.~\ref{alg:f-sim})}
    \State $\hspace{5mm}\mbox{f-sim}(\mathcal{D})\mapsto \{\mathcal{S}_r,\mathcal{S}_u\},\mathcal{M}_{A_{i,j}}^{n \times n} \hspace{39mm}\triangleright$ \emph{alg.~\ref{alg:f-sim} output}
    \State $\hspace{5mm}\mbox{textual-component:} \hspace{55.5mm} \triangleright$ \emph{invoke text-sim (alg.~\ref{alg:text-sim})}
    \State $\hspace{10mm}\forall v_i \in \mathcal{S}_r \hspace{2mm} \mbox{get k tweets} \hspace{46mm}\triangleright$ \emph{set of texts } $\mathcal{T}_{v_i}$
    \State $\hspace{10mm} \mbox{text-sim}(\mathcal{S}_r)\mapsto \{\mathcal{T}_r,\mathcal{T}_u\},\mathcal{M}_{ta}^{m\times m} \hspace{27.5mm}\triangleright$ \emph{alg.~\ref{alg:text-sim} output}
    \State $\hspace{10mm} \mbox{compare all topics}(\mathcal{T}_{v_i}\in \mathcal{S}_r) \mbox{ using Eq.~\ref{eq:js-divergence}}  \hspace{15mm}\triangleright$ \emph{affinity matrix}  
    \State local clusters:
    \State $\hspace{5mm} \psi(\mathcal{S}_r,\mathcal{T}_r) \hspace{69mm}\triangleright \mathcal{S}_r,\mathcal{T}_r \geq \tau$ 
    \State \textbf{Output:}
    \State $\hspace{5mm} \mathcal{C}_{c_i}^{m\times p} \hspace{77mm}\triangleright $ \emph{local communities}
        \end{algorithmic}
    \normalsize
    \end{algorithm}
% YK: Algorithm \ref{alg:mct} is not mentioned in text.
    
%In many applications, it is often the case that the dynamism of a system vis-\'{a}-vis some variables are analysed to understand how the system responds to changes, e.g. observing the greatest rate of increase or decrease at a given instance. The direction in which changes occurs defines the direction of the steepest ascent or descent \cite{strausscalculus2002}. 

\subsubsection{Optimising matrices of values}
\label{sec:optimising-matrices-of-values}

Recall that the set of textually related nodes $\mathcal{T}_r$ is a subset of the structurally related nodes $\mathcal{S}_r$ ($\mathcal{M}_{c_{vr}}$), i.e~$\mathcal{T}_r \subseteq \mathcal{S}_r$. 
Since the optimisation goal is to \textit{maximise} $\mathcal{T}_r$ ($\mathcal{M}_{vt}$), the two are equated under the constraint: $\mathcal{M}_{vt} = \mathcal{M}_{c_{vr}}$, such that $\mathcal{M}_{vt} - \mathcal{M}_{c_{vr}}= 0$. 
Noting the constraint, the simplified representation used in Eq.~\ref{eq:lagrangian-relaxation} also applies to $\mathcal{M}_{vt}$, given by $\mathcal{M}_{vt} = \mathcal{M}_{c_{vr}}=P$, to achieve the maximum values possible by determining the extremum of the function. 
Thus, the goal is to maximise the joint models under the constrained function according to Eq.~\ref{eq:mct-matrices-of-values-optimisation}\footnote{The proportionality constant $\lambda$ in Eq.~\ref{eq:mct-matrices-of-values-optimisation} denotes a Lagrange multiplier.}:
\begin{equation}
    \label{eq:mct-matrices-of-values-optimisation}
    \psi_{st}=||D-PQ^T||_F^2 + \sum_{i=1}^n\sum_{s=1}^k p_{is}\alpha _{is} + \sum_{j=1}^p\sum_{s=1}^k q_{js}\beta _{js} -  \lambda Tr(\mathcal{M}_{vt}^T\mathcal{M}_{va}\mathcal{M}_{vt})
\end{equation}
%The proportionality constant $\lambda$ in Eq.~\ref{eq:mct-matrices-of-values-optimisation} denotes a Lagrange multiplier. 
%See the Appendix, Section~\ref{appendix}, for details. 
% ALTERNATIVE VIEW ... MCT2:%MCT II
%,i.e.~ensuring that each node shares both $\mathcal{S}_r$ and $\mathcal{T}_r$ clusters where applicable. %, as illustrated in Figure~\ref{fig:strutural-content-clusters}). 

\subsubsection{Optimising intra-cluster similarity}%Intra-cluster similarity optimisation}
\label{sec:optimising-intra-cluster}%\paragraph{Nodes-reciprocal communities}

Intra-cluster similarity optimisation is similar to the approach in Section~\ref{sec:optimising-matrices-of-values} through the use of value matrices, but different objective function. 
The approach in Eq.~\ref{eq:lagrangian-relaxation} and the corresponding update rule (Eq.~\ref{eq:structural-update-rule}) are based on a matrix factorisation, which poses challenges with respect to exact or one-one mapping to the textually related clusters ($\mathcal{T}_r$). 
We know that the two are related at a higher level, since $\mathcal{T}_r \subset \mathcal{S}_r$, but the details about the shared clusters are not fully established. 
To address this challenge, we propose the following approach based on the node similarity. %\paragraph{Nodes-reciprocal communities} 
Information about similar nodes is stored in the nodes' affinity matrix ($\mathcal{M}_{va}^{n\times n}$), in which the magnitude of pairwise similarity decides entries in the matrix. 
Nodes are assigned to communities based on their degree of similarity denoted by $\mathcal{M}_{C_{vr}}^{n\times k}$ (of $n$ nodes and $k$ reciprocal-communities).  
For example, $\{c_{vr_1}, \dotsi, c_{vr_k}\} \in \mathcal{C}_{vr}$ represents a set of nodes-reciprocal communities and membership in a cluster is qualified by Eq.~\ref{eq:jaccard-index} - \ref{eq:probable-reciprocity}. 
With a higher probability of forming a tie for nodes in the same cluster, community detection is based on optimising the joint similarities of $\mathcal{S}_r$ and $\mathcal{T}_r$:
    \begin{equation}
        \label{eq:mct-intra-cluster-optimisation}
        \psi_{st}(v_i, v_j) = (\lambda)\cdot\mathcal{S}_r(v_i,v_j) + (1-\lambda)\cdot\mathcal{T}_r(v_i,v_j)
    \end{equation}
The goal of Eq.~\ref{eq:mct-intra-cluster-optimisation} is to maximise the joint similarity between $\mathcal{S}_r$ and $\mathcal{T}_r$ according to an aggregation criterion inspired by \cite{aggarwal2012event}, based on the similarity scores between pairs and a user-defined balancing parameter\footnote{Note that this is different from the one used in optimisation based on matrices of values.} $\lambda$, with values in $(0,1)$. 
We follow the approach in \cite{prokhorenkova2019using} to find optimum value for the $\lambda$. Algorithm~\ref{alg:mct-intra-cluster-algorithm} describes how nodes are assigned to relevant clusters until the stopping criterion, a user-defined integer $M$ signifying the desired number of clusters, is reached. 
%%%%%%###MORE UPDATES HERE .... 

%##############%Algorithm~\ref{alg:mct-intra-cluster-algorithm} describes the implementation process. 
    \begin{algorithm}[t]
    \small
    \caption{\emph{: Algorithm MCT-2} identifies local communities known as \textit{microcosms} in a network.}
    %\label{alg:mct-2}
    \label{alg:mct-intra-cluster-algorithm}
        \begin{algorithmic}[1]
        \State \textbf{Input:} a collection of network data $\mathcal{D}$
        \State structural-component:
        \State $\hspace{5mm}\mbox{f-sim}(\mathcal{D})\mapsto \{\mathcal{S}_r,\mathcal{S}_u\},\mathcal{M}_{sa}^{n\times n}$ 
        \State $\hspace{5mm}\mbox{textual-component:} \hspace{55mm}$ %\triangleright$ 
        \State $\hspace{10mm}\forall v_i \in \mathcal{S}_r \hspace{2mm} \mbox{get k tweets}$ 
        \State $\hspace{10mm} \mbox{text-sim}(\mathcal{S}_r)\mapsto \{\mathcal{T}_r,\mathcal{T}_u\},\mathcal{M}_{ta}^{n\times n}$ 
        \State $\hspace{10mm} \mbox{compare all topics}(\mathcal{T}_{v_i}\in \mathcal{S}_r) \mbox{ using Eq.~\ref{eq:js-divergence}}$ 
        \State \textbf{Clusters initialisation:} 
        \State $\hspace{5mm} \mbox{select four random seed nodes: } v_i,v_j,v_k,v_l \in \mathcal{S}_r$ \mbox{ and } $v_i,v_j,v_k,v_l \in \mathcal{T}_r$ 
        \State $\hspace{5mm} \mbox{compute pairwise similarities among } v_i,v_j,v_k,v_l \mbox{ using } \psi_{s_rt_r}(v_i, v_j)$ 
        \If{$\mathcal{T}_{sim}(\mathcal{T}_{v_i}, \mathcal{T}_{v_j}) \geq \tau$} 
        \State \emph{create single cluster} $\mathcal{C}_{ij}$
        \Else
        \State \emph{create two clusters} $\mathcal{C}_{i}, \mathcal{C}_{j}$
        \EndIf
        \State $\mbox{repeat 9 -- 15 until} |\mathcal{C}_{ik}|_{k=1}^{M} = M \hspace{20mm}\triangleright$\emph{maximum clusters M}
        \State \textbf{Assign nodes to clusters:} 
        \State $\hspace{5mm} \forall v_i \in \mathcal{S}_r \mbox{ compute similarity with cluster's mean}$ 
        \State $\hspace{5mm} max_{\phi(v_i, \mu _{C_i})} \hspace{45mm}\triangleright$ \emph{assign $v_i$ to the most similar $\mu _{C_i}$}
        \State \emph{update cluster's mean:}  $\mu _{C_i} \longleftarrow  \mu _{C_i}$
        \State \textbf{Output:}
        \State $\hspace{5mm} \mbox{local communities}$
        \end{algorithmic}
    \normalsize
    \end{algorithm}
%##############
%The rank, $r$ of the matrix should be chosen so that $(n+p)r < np$ \cite{lee2001algorithms}. ... the selection criteria put forward by the authors.

%The Frobenius Norm is used to assess or quantify the quality of the approximation. Through multiplicative update rule, the cost function optimised and the algorithm (NMF) is guaranteed to reach/converge to a local minimum in the function \cite{lee2001algorithms}. %Achiveing global minimum is not possible because the function is convex only in one of the approximation parameters -- P or Q -- not both \cite{lee2001algrithms} ... 

%On Algorithms for evaluation For the sake of evaluation, we group the relevant algorithms according to the following: Algorithms purely based on structural aspect (S-based); Algorithms purely based on content/textual aspect (T-based) ... any relevant algorithms? and Algorithms based on both structural and textual aspects (S&T)

%% file: p4-experimentation.tex
\section{Experimentation}
\label{sec:experimentation}

This section presents our experimentation to evaluate the MCT against other existing methods.

\subsection{Datasets}
\label{sec:mct-datasets}

We utilise the following diverse datasets for the experimentation. %We employ empirical and ground-truth datasets from a public repository.
% , predicted and Ego-networks,

\subsubsection{Ground-truth and predicted data} 
%Because a collection of reciprocal ties is regarded as a facilitator for the detection of socially cohesive communities \cite{inuwa2019simmelian}, we present a useful strategy to maximise the use of nodes with reciprocal ties for mining tasks involving Twitter. 
Unlike previous studies in which datasets from various social networks were collected \citep{leskovec2012learning,yoshida2013toward,yang2015defining}, this study focuses on nodes with reciprocal, not directed, ties. The reciprocal collection consists of \textit{dyadic} and \textit{transitive} datasets, which were collected using Twitter's Application Programming Interface (API) according to Algorithm~\ref{alg:search-dyads}. 
The process returns a collection of \textit{tweet objects}, a complex object with many descriptive fields, which allows to extract structural and textual components for analysis. 
The collection begins with a search on the network profile of each from a finite set of seed users\footnote{Seed users are verified or unverified accounts devoid of spammers or social bots were collected by SPD filtering \citep{inuwa2018detection}. A \textit{'list'} on Twitter allows a user to store a set of preferred users and can be used to obtain relevant information.}, or a network composition $m_{v_i}$, consisting of lists of friends $fr_{v_i}$ and followers $fl_{v_i}$, to determine user pairs that follow each other. 
The set of reciprocal pairs is denoted by $\kappa \in m_{v_i}$ and the transitive dataset is a scaled-version of dyadic data. 

In addition to the collection of nodes with actual pairwise ties (denoted as G-pTie in Table~\ref{sec:mct-datasets}), the ground-truth dataset also consists of public data associated with COVID-19 outbreak (G-pMention) related to aspects of scepticism and myths about the pandemic \cite{inuwa2020curated}. 
The data contains two broad categories: information put forward by credible sources, such as the World Health Organisation (WHO), and information from users dismissing WHO's guidelines on combating the pandemic. 
The dataset consists of interaction information about users who mention each other. Nodes with frequent mentioning are highly likely to be in the same community. 
For the dataset consisting of predicted pairwise ties (P-pTie), a reciprocal tie exists between $v_i$ and $v_j$ if $p(R_{v_i,v_j})\geq \tau$, otherwise just a directed tie. In Table~\ref{tab:microcosm-detection-dataset}, SND1 refers to synthetic network data generated based on LFR approach (see Section~\ref{sec:baseline-models} for details).

    \begin{algorithm}[t]
    \small
    \caption{\emph{Algorithm search-dyads} profiles users with directed and undirected ties on Twitter}
    \label{alg:search-dyads}
        \begin{algorithmic}[1]
        \State \textbf{Initialisation:} $1-edge\longrightarrow \{\}, dyads \longrightarrow \{\}$
        \State \textbf{Input:} begin with an arbitrary set of seed users, say $k$
        \While {$k \neq \emptyset$}
            \State $\forall v_i \in k$, get sets of \textit{friends} $fr_{v_i}$, \textit{followers} $fl_{v_i}$, $\hspace{3mm}$ $fr_{v_i},fl_{v_i} \in m_{v_i}$; $m_{v_i}$ denotes $v_i$ network 
            \State $\forall v_j \in fr_{v_i}$, retrieve the sets $fr_{v_j}$ and $fl_{v_j}$, $\hspace{3mm}$ $fr_{v_j},fl_{v_j} \in m'_{v_j}$; $m'_{v_j}$ denotes $v_j$ network
        \If{$v_i \in fr_{v_j}$}
            \State $v_i \sim v_j$ \hspace{10mm} $\triangleright$ both follows one another
            \State \emph{update dyads}
        \Else
            \State $v_i$ follows $v_j$
            \State \emph{update 1-edge}
        \EndIf
        \EndWhile
        \end{algorithmic}
    \normalsize
    \end{algorithm}

 \begin{table}[!b]
          \footnotesize
          \caption[Microcosm detection datasets summary]{A summary of microcosms detection datasets. V and E denote the node and edge size, respectively. G-pTie and P-pTie denote groundtruth and predicted sets of users with pairwise connectivity; G-Mention denotes collection of users with pairwise mentioning; $\mu_{deg.}$ refers to the average degree in each data category.} 
          \label{tab:microcosm-detection-dataset}
          \begin{tabular}{l>{\raggedright}p{3.0cm}cclp{3cm}} \hline %ccp{3cm}} \hline
            & Category &  V & E & Description  \\ \hline 
            \multirow{8}{*}{\rotatebox{90}{\textbf{Datasets}}} 
            & G-pTie & 18973 & 15538 & $\mu_{deg.} =  1.6379$ \\ %first row
            & P-pTie & 15038 & 1298998 & $\mu_{deg.} = 172.7621$ \\ %second row
            & G-pMention & 514 & 259 &  $\mu_{deg.} = 1.0078 $ \\%second row 
            & Karate club & 34 & 78 & Consists of 2 groundtruth communities \\
            & Dolphin & 62 & --- & Consists of 2 groundtruth communities \\
            & Pol. Blog & 1224 & --- & Consists of 2 groundtruth communities \\ 
            & ego-Facebook & 4039 & 88234 & $\mu_{deg.} = 43.6910$ \\ 
            & ego-Twitter & 81,306 & 1768149 & --- \\
            %& Synthetic & --- & --- & --- & --- & --- & ---\\ \hline
            & SND1 & 1000 & --- & ---  \\ \hline
          \hline
        \end{tabular}
        \normalsize
        \end{table}

%Where possible, we need to approximate each dataset to the nearest hundred/thousand ..... and rerun

\subsubsection{Public datasets}
\label{sec:public-datasets}
To reinforce evaluation and generalisation, we use the following collections of publicly available datasets. %\cite{snapnets}. 
The datasets consist of real-world networks commonly used for community detection. Essentially, the following datasets have been used: Zachary’s karate club \cite{zachary1977information}, dolphin social network \cite{lusseau2003bottlenose}, political blog dataset \cite{adamic2005political} and Ego-network, consisting Facebook and Twitter datasets \cite{snapnets}. %%These are datasets with ground-truth labels. 
The Facebook data contains anonymised \textit{'circles'} or \textit{'friends lists'}, and  \textit{node features (profiles)}. 
Each node has \textit{node ids}, sets of \textit{connections} or \textit{edges}, and \textit{anonymised features} encoding information about its \textit{circle}. The Facebook data allows to explore communities using each user's network circle in terms of size and diversity of membership.
Moreover, the collection consists of \textit{synthetic data}, which is based on the approach proposed in \cite{prokhorenkova2019using} to generate synthetic network with known parameters. The synthetic nature of the networks makes it possible to explore the parameters' space for the best community structure in the network. 
Table~\ref{tab:microcosm-detection-dataset} shows basic statistics of datasets used in this study. 
%The dataset table shows relevant parameters in the synthesised network and the predicted version ... and some evaluation metrics. 
%The table also shows extra information, such as \textit{closeness centrality}, \textit{betweenness} and \textit{clustering coefficient}, for each dataset.  

\begin{figure}[!b]
    \centering
    \includegraphics[width=0.95\linewidth]{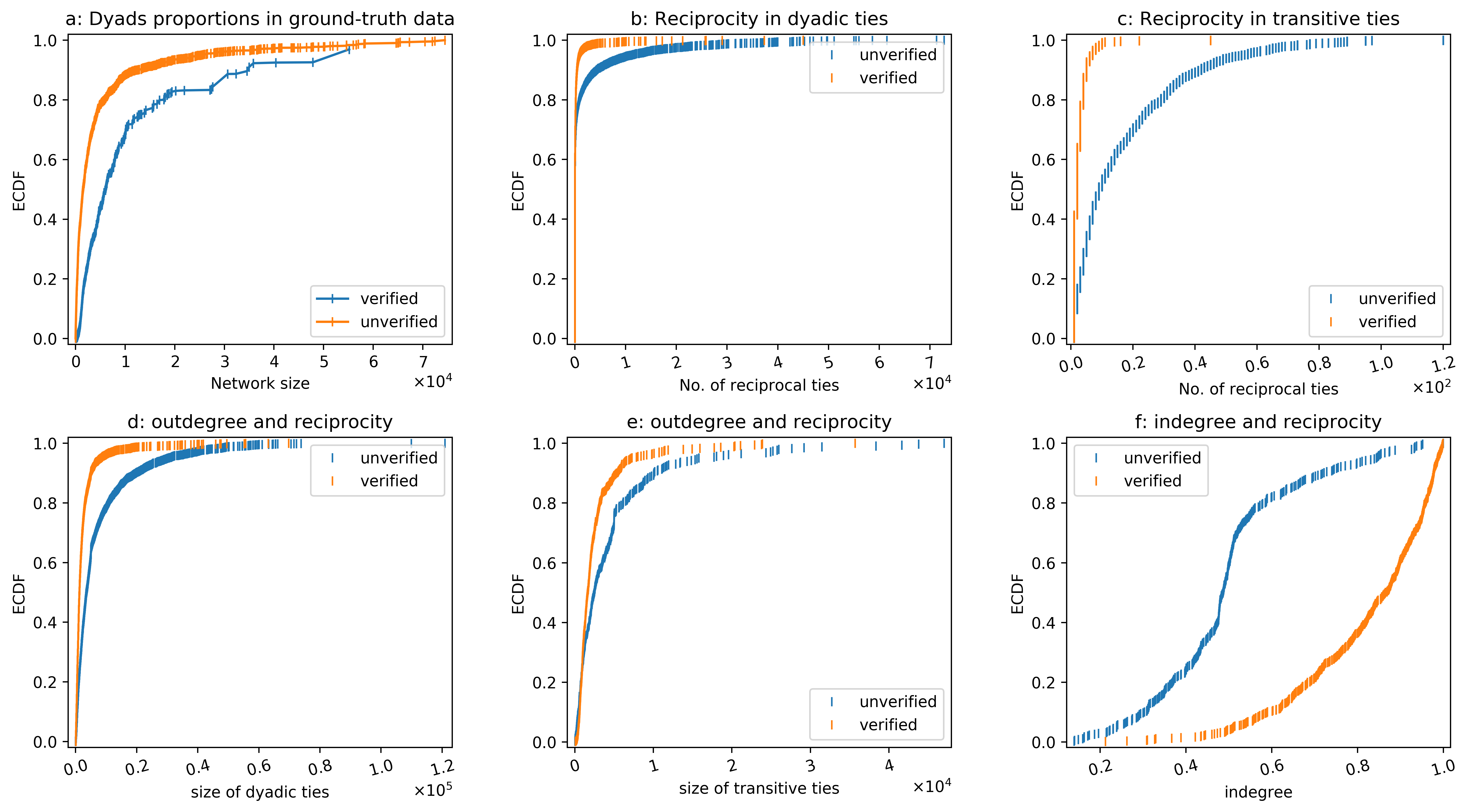}
    \caption[ECDF of reciprocal ties]{Sub-figures (a), (b) and (c) show the reciprocity effect on nodes with many dyadic relationships. Sub-figures (d), (e) and (f) show outdegree-reciprocity and indegree-reciprocity relationship in the ground-truth data.}
    \label{fig:ties-ecdf}
\end{figure}

\subsection{Meta-analysis}
\label{sec:meta-analysis}

Owing to the prevalence of unreciprocated and event-type ties on Twitter, we conjecture that mining tasks, such as community detection, are less effective and more challenging. 
In this section, our goal is to apply a pragmatic approach that provides a statistical analysis of relevant metrics in the datasets to identify strongly correlated node attributes (Figure~\ref{fig:tie-formation-and-strength}(b)) with reciprocity among nodes. 
The empirical cumulative distribution function (ECDF) gives the probability of a quantity evaluated at arbitrary points.
We use it to analyse observations, such as the variation of dyads or Simmelian ties across user categories or network size.

\subsubsection{Proportion of reciprocal units}
\label{sec:reciprocal-units}

\begin{figure}[t]%[tbhp!]
          \centering
          \includegraphics[width=0.95\linewidth]{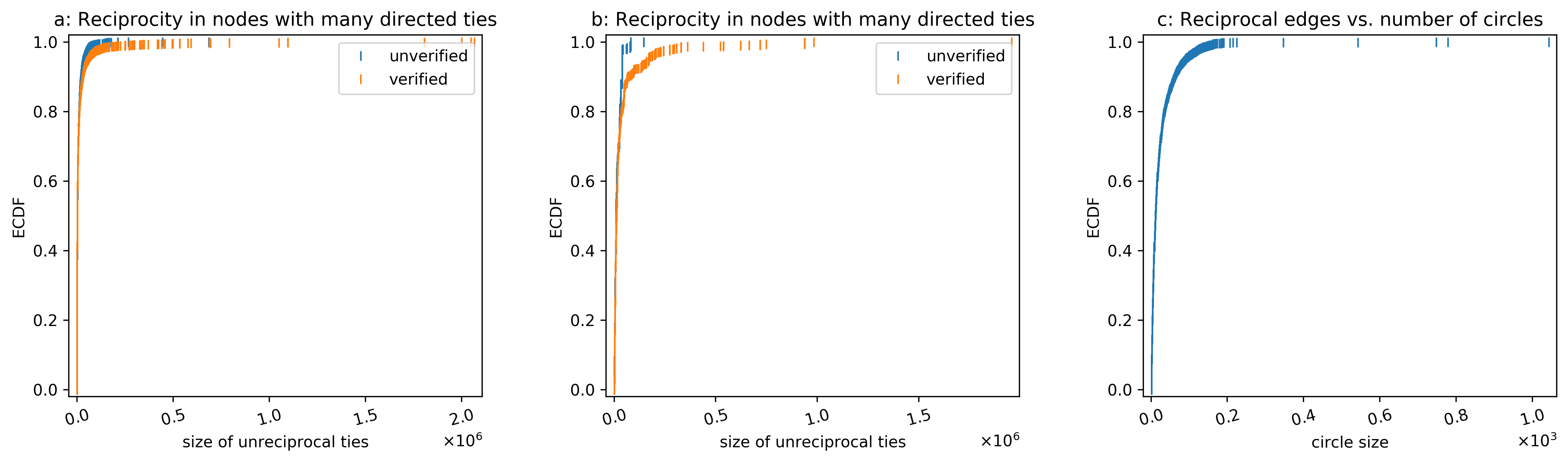}
          \caption[Reciprocity effect of features]{Relationship between the number of directed ties and reciprocity across different nodes in the datasets.}
          \label{fig:reciprocity-effect-and-features}
     \end{figure}

Noting the flexibility of connections and the rarity of reciprocal links on Twitter, large scale dyadic ties are rare and difficult to locate. 
Using Algorithm~\ref{alg:search-dyads}, we collected \textit{directed} or \textit{1-edge} and \textit{undirected} data, and examined the \textit{network topology} of each category and its utility in the detection of local communities. In
Figure~\ref{fig:ties-ecdf}, there is a high proportion of reciprocity in unverified users in comparison to the verified counterpart. 
The reciprocity ability slightly decreases with increasing network size of the user, which can be attributed to the difficulty in keeping track of and responding to all followership requests. 
Sub-figures~\ref{fig:ties-ecdf}(a) and (b) show the relationship between reciprocity and the number of reciprocal ties. 
While there is higher reciprocity in the unverified category, the verified category shows almost 100\% reciprocity with a relatively small network size. 
Sub-figure~\ref{fig:ties-ecdf}(c) shows similar behaviour in transitive ties, but is more evident in the unverified users category. 
Similarly, sub-figures~\ref{fig:ties-ecdf}(d),(e) and (f) show the relationship between outdegree and reciprocity and indegree and reciprocity in the ground-truth data. 
The behaviour resembles an inverse relationship in which reciprocity decreases with increasing outdegree (sub-figures~\ref{fig:ties-ecdf}(d) and (e)). 
Sub-figure~\ref{fig:ties-ecdf}(f) shows an almost linear relationship between indegree and reciprocity, especially among the unverified category. 
In the verified category, the effect is low and seems to shoot once the network size increases (vis-a-vis indegree/number of followers). 
There is an instant reciprocity in the unverified users, which can be explained by suggesting that the users are interested in expanding their network. 
Figure~\ref{fig:reciprocity-effect-and-features} shows the relationship between the number of directed ties and reciprocity across user categories in the data. 
The results demonstrate that verified users have many directed nodes or unreciprocated ties but with less reciprocity. 
This observation holds for nodes with many dyadic and transitive ties in the data. 

\subsection{Evaluation}
\label{sec:mct-evaluation}
To ascertain the efficacy and relevance of the study's output, evaluation entails thorough analyses and comparison with relevant baselines drawn from the literature. 
Quantitative analyses of experimentation on various datasets using the baseline algorithms is involved. 
Other forms of evaluation are specific to the structural and textual levels of the MCT strategy. 
The evaluation process aims to: 
(a) investigate the effect of utilising structurally-related nodes in identifying local communities in social networks, 
(b) compare structurally-related clusters with textually-related clusters, and 
(c) evaluate the performance of MCT in comparison with baseline models. 

\subsubsection{Evaluation metrics}
\label{sec:evaluation-metrics}

This section discusses quantitative measures for validating the performance of MCT and baseline methods. 
Because of the multilevel approach, the metrics are suitable for evaluating network structure (structural clusters) and textual-clusters (roughly considered as labels).

\paragraph{Clustering coefficient and Community cohesion}
Clustering coefficient, $C_{coeff}$), is used to quantify the clustering tendency of a given node in relation to other nodes within a network \cite{watts2007influentials}. Computing $C_{coeff}$ requires: $edges = \frac{k_i(k_i-1)}{2}$ and $C_{coeff_i} = \frac{2E_i}{k_i(k_i-1)}$ where $i,k_i, E_i$ denote a network node, the number of edges connecting $i$ to $k_i$ other nodes in the network, and the actual number of existing edges between $k_i$ nodes, respectively. The ratio $E_i \propto \frac{k_i(k_i-1)}{2}$ defines the clustering coefficient of a node. \textit{Community cohesion} demonstrates the level of connectivity within a community and is captured by measuring the degree of cohesiveness. 
Due to the presence of a strong connectivity among nodes, a well-connected community is intuitively difficult to divide into sub-communities \cite{leskovec2010empirical}. 
Any useful metric that reveals the degree of cohesion can be used to evaluate cohesiveness, i.e., if the community is well-connected and difficult to partition. In this study, cohesiveness is measured by the degree of similarities $\mathcal{S}_r$ and $\mathcal{T}_r$. We compute the \textit{average degree ($\mu _{deg}$)}, defined as the average node degree to other member nodes \cite{radicchi2004defining}. Moreover, we use the \textit{accuracy metric}, i.e., the fraction of predicted labels to the total number of data points. 

\paragraph{Modularity and NMI} Modularity, $\mathbf{Q}$, measures the strength of communities, as the number of edges falling within groups minus the expected number in an equivalent network with edges placed at random \cite{newman2004fast}. 
Usually, $\mathbf{Q} > 0$ signifies the possible presence of a community structure and the higher the values the better \cite{newman2004detecting}. Normalised Mutual Information (NMI) is another statistical tool to evaluate the quality of network clusters \cite{danon2005comparing}. 
NMI measures the degree of agreement between network partitions, based on the assumption that each node in a community, $v_i \in \mathcal{V}$, is associated with both the \textit{true community} and the \textit{predicted community}, such that $l_{v,p} = i$ defines the predicted community $i$ of a node \cite{fred2002data}. Furthermore, we apply \textit{Rand} and \textit{Jaccard} similarity metrics, which are based on tracking both correctly and incorrectly classified pairs of nodes, especially in groundtruth datasets. 
%Eq.\ref{eq:nmi-nmi} is used to compute NMI: 
    %\begin{equation}
     %   \label{eq:nmi-nmi}
     %   NMI = \frac{-2\sum_{i=1}^{k_t}\sum_{j=1}^{k_p}n_{i,j}^{tp}\log (\frac{n_{i,j}^{t_p}\cdot n}{n_i^t \cdot n_j^p})}{\sum_{i=1}^{k_t}n_i^t \log(\frac{n_i^t}{n})+ \sum_{j=1}^{k_p}n_j^p\log(\frac{n_j^p}{n})}
    %\end{equation}

\subsubsection{Baseline Models}
\label{sec:baseline-models}

For evaluation, \textit{MCT} is applied alongside the following detection algorithms with different modes of operation on the datasets described in Section~\ref{sec:mct-datasets} to identify local community structures. %ALGORITHMS BASED ON STRUCTURAL ASPECT

\paragraph{Girvan-Neuman (G-N) and Label propagation (LP)} 
The G-N algorithm assumes that a community detection algorithm can naturally detect divisions among vertices without external influence or imposed restrictions on the divisions \cite{newman2004finding}. 
Accordingly, \citet{girvan2002community} proposed the iterative G-N algorithm that progressively removes network edges based on betweenness, a  metric to quantify traffic flow among nodes. 
Each node's betweenness score dictates which edge to remove. 
The most critical nodes are likely to experience high traffic flow, hence will possibly create a bottleneck. 
%Ultimately, the \textit{G-N algorithm} traces and discards those nodes thereby achieving natural divisions in the network.
%\paragraph{Label propagation (LP)} 
The LP algorithm is an iterative clustering method that converts unlabelled data to labelled given an initial seed of labelled data. 
Labelling involves a repetitive random node reshuffling and tagging with the most frequent label among its neighbours until convergence \cite{zhu2002learning}. 
The labelled data information is then propagated across the whole network.

\paragraph{Synthetic Network Model}
This is achieved using the widely used approach, or LFR model, proposed in \cite{lancichinetti2008benchmark} to generate synthetic networks with planted partitions or community structures. For a given network $G$ generated via the LFR, the following basic model's parameters are defined: $\gamma, \beta, \bar{d}, \hat{\mu}$ denoting exponents of the power-law degree distribution, community size distribution, mean degree and mixing parameter, respectively. Accordingly, the model ensures that nodes' degrees are sampled independently whose distribution exhibit power-law behaviour and the mixing parameter, $\hat{\mu}$, to distribute nodes' indegree and outdegree such that $1-\hat{\mu}$ and $\hat{\mu}$ denote the proportions of edges shared with nodes in the same and different communities, respectively. 
The SND1 netwrok in Table~\ref{tab:microcosm-detection-dataset} is generated based on the LFR approach. 
The network consists of 1000 nodes, $\gamma = 1.5$, $\bar{d} = 15$, $\gamma_C = 0.8$, $C_{min} = 30$, $C_{max} = 300$, and the mixing parameter ,$\hat{\mu}$, sampled from ${0.1, 0.01, 0.2, 0.3, 0.5, 0.7, 0.9, 1}$. Because the parameters pertaining the network and the embedded community structure are known, relevant community detection methods should be able to detect or identify values (especially for the community) that approximate such parameters.
    
%\paragraph{Planted Partition Model (PPM)}
The Planted Partition Model (PPM) is a form of likelihood optimisation algorithms that are commonly used for community detection task. Due to their mathematical efficacy, many algorithms are defined based on relevant assumptions about the underlaying structure in the network. Under this approach, a network is a composition of communities, which are used to infer the network \cite{bickel2009nonparametric}. %... this is roughly expressed as conditional probability, $P(G|C$ where G is the network/graph and C the set of its communities. Nodes are assigned to communities, e.g. probabilistically, and the best set of communities representative of the network is returned. % via the shared community
The PPM relies on the community membership of nodes to probabilistically decide whether any pairs of nodes are connected. We apply variant of the PPM (degree-corrected planted partition model \cite{karrer2011stochastic}) and an extended version of the LFR model proposed in \cite{prokhorenkova2019using} as part of the evaluation.

\subsection{Detection of community structure}
\label{sec:how-good-is-f-sim}

In this section, we focus on the detection of community structures using our proposed method, introduced in Section~\ref{sec:microcosm-detection-algorithm}, and the baseline models, described in Section~\ref{sec:baseline-models}. 
The detection process consist of four steps: 
(1) retrieve a set of nodes with reciprocal ties on Twitter, 
(2) compute the similarity proportion between pairs, using Algorithm~\ref{alg:f-sim}, 
(3) compare prediction accuracy using the ground-truth, and  
(4) perform clustering for community detection. 

\subsubsection{Effectiveness of tie prediction}
Using Algorithm~\ref{alg:f-sim}, which computes the similarity between the corresponding features of any pairs of nodes, we report its efficacy in the prediction pipeline. 
Due to the availability of empirical data, the effectiveness of the model is quantified with respect to the degree of conformity with the ground-truth data. 
This is vital because the tie prediction segment is not relevant if it does not add value to the overall detection framework. 
The accuracy of the prediction is obtained by computing the ratio of predicted reciprocal ties to true reciprocal ties. The best result achieved is $.608$ accuracy; depending on the threshold $\tau$, the accuracy may be lower or higher. Sub-figure~\ref{fig:line_ecdf_and_fsim_plots}(c) shows possible values of $\tau$ and the corresponding accuracy. %Moreover, we focus is on the effect of using a collection of nodes with structural relationships in identifying local communities in social networks. The investigation begins with sets of structurally-related and structurally-unrelated nodes. 

\subsubsection{Community structure} 
We examine how the use of a collection of structurally-related nodes affects community detection, and compare performance. 
For the experiments, we apply the proposed method, (MCT), and the baselines, G-N \cite{girvan2002community} and LP \cite{zhu2002learning}. 
Table~\ref{tab:mct-and-baselines-clustering-performance} shows the results of applying the community detection algorithms on the data according to the evaluation metrics described in Section~\ref{sec:evaluation-metrics}. 
Although all the algorithms detected community structures, there are quantitative variations among the outcomes. 
Our analysis is along the following dimensions.

\begin{figure}[t]
    \centering
    \includegraphics[width=\linewidth]{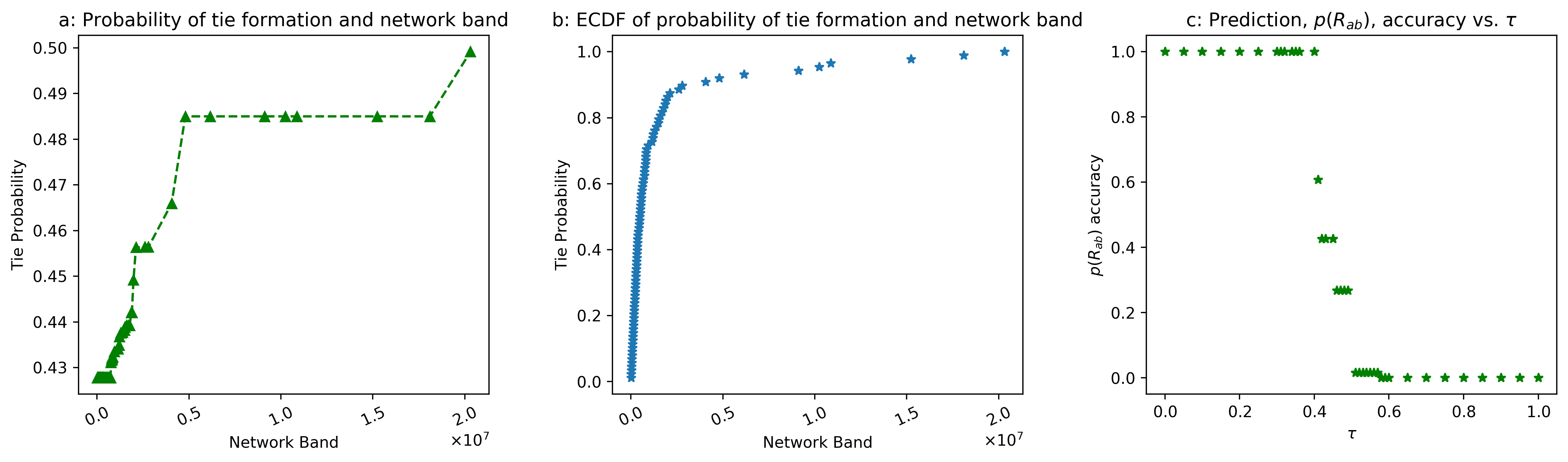}
    \caption[Line and ECDF plot of probable tie formation]{Sub-figures $a$ and $b$ show the probability of tie formation as a function of \textit{network size}; there is a high chance of reciprocating a tie among users in a network band of $0.5\times 10^7$. Sub-figure $c$ shows the prediction accuracy versus the threshold value in \textit{f-sim} (Algorithm~\ref{alg:f-sim}). The prediction accuracy is almost 100\% when the value of the threshold is low; conversely, the accuracy is almost 0\% when the threshold value is very high. A \textit{switch-point} can be observed toward the midpoint in which the accuracy is just above 60\% at a threshold value of about 0.41. With additional features, the prediction can be improved. For instance, the inclusion of a description feature led to a significant improvement; however, it requires training on a large corpus to obtain the embedding of terms in text.}
        \label{fig:line_ecdf_and_fsim_plots}%\label{fig:line_and_ecdf_plots_of_ties} and \label{fig:prediction-accuracy-and-threshold}
\end{figure}
    
\noindent \emph{Effect of datasets:} 
All algorithms perform best on the ground-truth data, followed by \textit{ego-Facebook}, then \textit{predicted}, and worst on \textit{ego-Twitter}. 
The \textit{ego-Facebook} data consists of nodes with reciprocal ties, but the textual feature set is small, making it less complex than the other datasets. %(as described in Table~\ref{tab:fb-data}) 
We consider \textit{homophily} and \textit{structural equivalence} as precursors of communities, in which nodes with similar profiles or social status are more likely to interact and establish a small community. 
For instance, sub-figures \ref{fig:line_ecdf_and_fsim_plots}($a$) and ($b$) show homophily as a form of structural equivalence based on network size and indegree for examining the probability the formation of an edge. 
Sub-figures \ref{fig:line_ecdf_and_fsim_plots}($a$) and ($b$), depicts a behaviour that resembles an inverse relationship: increase in network size results in decrease in reciprocity. 

\begin{table}[!b]
    \centering
    \footnotesize
    \caption[Clustering performance]{Results of experiments on three datasets for community detection using algorithms based on structural properties. G-N: Girvan--Neuman, LP: Label Propagation, MCT: Multilevel Clustering Technique, \#DC: Number of Detected Communities}
    \label{tab:mct-and-baselines-clustering-performance}
    \begin{tabular}{lccccccccc}  \hline
        & \multicolumn{3}{c}{G--N} & \multicolumn{3}{c}{LP} & \multicolumn{3}{c}{MCT}\\ %first row
        Dataset & \multicolumn{2}{c}{Metric} &\#DC &\multicolumn{2}{c}{Metric} &\#DC &\multicolumn{2}{c}{Metric} &\#DC \\ %second row (inner)
        & Q & NMI  &  & Q & NMI & & Q & NMI &  \\%third row (most inner) 
        \hline
        G-pTie &.908& .794 & 308 &.77& .602 & 1319 & .915 & .791 & 263 \\
        ego-Twitter &.334& .197 & 1431 &.215& .131 & 2131 &.307 & .230 & 1131 \\
        ego-Facebook  &.522 & .590 & 1037 &.421 & .304 & 1780 &.503 & .372 & 1845 \\
        P-pTie & .473 & .311 & 1107 & .360 & .267 & 2071 &.601 & .472 & 985 \\
        \hline
    \end{tabular}
    \normalsize
\end{table}

\noindent \emph{Effect of models:} Table~\ref{tab:mct-and-baselines-clustering-performance} also demonstrates the performance of each model. The \textit{MCT} results indicate a more localised structure noting the magnitude of \textit{Q}, \textit{NMI} and the number of detected communities (\textit{\#DC}) with respect to the \textit{ground-truth} data. 
We attribute the improvement to the use of in-depth structural features that introduce a connectivity layer. 
\textit{MCT} explores the data for community structures at local and global level through a high-level grouping of nodes into communities, according to the network size and the recognition of bi-modal information sources. 
%The following result is based on the formulation of the MCT-II to detect community structure. The formulation/implementation requires a learning parameter or hyperparameter to balance the equation given in Eq.~\ref{}. For that, we use the baseline models and the datasets describe in Section~\ref{}

For the parameterised approach, groundtruth datasets form the basis of the evaluation. Therefore, we discuss the results obtained in Table~\ref{tab:mct-and-baselines-clustering-performance2} using Rand and Jaccard scores as the evaluation metrics. Generally, the results in the Table indicate good performance, especially with respect to the Rand score and on small datasets such as the Karate club. The values associated with the MCT signify better performance across all the datasets. However, there are instances where the algorithm's performance lags behind. For instance, the PPM and ILFR perform better on the Karate and Pol. Blogs datasets, which we attribute to the small size nature of the datasets. Moreover, there is a significant improvement on performance on the synthetic datasets, i.e. SND1. This s expected since the network consists of well defined community structure. A common trait among the algorithms is that they perform poorly based on the Jaccard index, suggesting that the metric is somewhat strict or further optimisation is needed. 

%TABLE 2: NMI .... 
    \begin{table}[!b]
        \centering
        \footnotesize
        \caption[Clustering performance]{Results of experiments on three datasets for community detection using relevant algorithms and evaluation metrics.}
        \label{tab:mct-and-baselines-clustering-performance2}
        \begin{tabular}{cccccccccccc}  \hline
            & \textbf{Model}  & \multicolumn{2}{c}{PPM} & \multicolumn{2}{c}{LFR} & \multicolumn{2}{c}{MCT} & \multicolumn{2}{c}{ILFR} \\ %first row
            %  \multicolumn{1}{c}{} &\multicolumn{2}{c}{} &\multicolumn{2}{c}{}
            %      & \multicolumn{2}{c}{} &\multicolumn{2}{c}{} &\multicolumn{2}{c}{} \\ %second row (inner) 
            & \textbf{Metric} & Rand & J & Rand & J & Rand & J & Rand & J \\%third row (most inner) 
            \hline
            \multirow{9}{*}{\rotatebox{90}{\textbf{Datasets}}} &
            G-pMention & 0.501 & 0.005 & 0.501 & 0.005 & 0.662 & 0.169 & 0.501 & 0.005 \\ 
            & Karate club & 0.734 & 0.475 & 0.734 & 0.475 & 0.701 & 0.461 & 0.720 & 0.425 \\ 
            & Dolphin & 0.651 & 0.379 & 0.581 & 0.255 & 0.696 & 0.340 & 0.536 & 0.167 \\ 
            & Pol. Blob & 0.903 & 0.821 & 0.878 & 0.775 & 0.565 & 0.487 & 0.581 & 0.166 \\ 
            & G-pTie & 0.621 & 0.012 & 0.621 & 0.012 & 0.629 & 0.210 & 0.621 & 0.012 \\ 
            & P-pTie & 0.652 & 0.001 & 0.652 & 0.001 & 0.696 & 0.237 & 0.652 & 0.001 \\ 
            & SND1 & 0.846 & 0.431 & 0.795 & 0.395 & 0.879 & 0.510 & 0.601 & 0.317 \\
            & ego-Facebook & 0.683 & 0.317 & 0.597 & 0.301 & 0.697 & 0.332 & 0.579 & 0.298 \\  
            \hline
        \end{tabular}
        \normalsize
    \end{table} 
    
%When the network size is huge, it is challenging to authoritatively specify when a given community detection algorithm will converge. Thus, we rely on a single iteration to analyse the algorithm's complexity, which will provide insights to its future performance. 
%Execution wise, the complexity of the MCT is relatively low. However, it tends to increas with growing datapoints. Let us assume that the execution complexity of a basic parametrised algorithm is f(C), then the term O(f(C) * s * r * m), where s is the number of comparisons in deciding the next cluster, s is the size size and m the number runs. The complexity of the algorithm needs further improvement. 
%Using the MCT strategy we obtaain significant improvements over the baselines. The underlying idea is to explore and utilise textual and structural aspects in uncovering community structure at various levels of granularity. 

%% file: p5-conclusion.tex
\section{Discussion}
\label{sec:discussion}

\noindent In this section, we discuss some significant observations from the study. 

\paragraph{Impact of reciprocal units and text aggregation for clustering}
One of the assumptions of this is that recognising a set of reciprocal units for community detection offers a more cohesive community representation. 
Since small groups allow modular analysis of social networks \cite{freeman1996some, dunbar1998social}, we examined reciprocal ties, dyadic and Simmelian, as the basic units of relational interaction on Twitter. 
However, Twitter's flexible and eccentric connections entangle locating nodes with reciprocal links. 
Structural similarity allows to organise nodes into connected clusters and simplifying community detection. 
Structurally similar nodes are more likely to connect and belong to the same community. 
%Using datasets with a large proportion of reciprocal units enhances the ability to recognise more cohesive groups.
%\subsection{The use of aggregate text for clustering}
%\subsection{Impact of using aggregate text for clustering} 
%
The high volume and small size of tweets make comparisons of discussions context challenging. 
Because a single tweet may not yield enough information about a discussion, we need to balance between quantity and quality. 
We collected a finite set of tweets from each node $v_i$ that defines a user corpus $\mathcal{T}_{v_i}$, and computed its overall theme to compare with other nodes. 
Textually-related nodes $\mathcal{T}_r$ are identified by a topic modelling technique that compares the similarities of the discussion topics of structurally similar nodes. 
%Moreover, analysing the textual content will provide a means of studying constructionism, a sociological premise that people who share knowledge are more likely to interact and is regarded as a means of forming social ties \cite{carley1991theory}, on Twitter. 

%\subsection{Improving social cohesion in the detection task}
\paragraph{Improving social cohesion in the detection task}%Utility of the MCT framework}
Online content increases rapidly in volume and complexity and is dominated by influential users.  These facts make the detection of socially cohesive groups on Twitter challenging. 
% This limitation hinders the full realisation of the benefits in communities such as cliquishness and local coherence. 
With respect to \textit{sociometry}, the formation of a social tie can be based on \textit{event-type} or \textit{state-type ties}. 
The size of a network and the size of communities are almost linearly correlated. 
Similarly, the size of a network is inversely correlated with its degree of homogeneity. 
The degree of interaction is higher among structurally similar users. 
Often, users that discuss with and mention each other are engaged in reciprocal ties, showing strong social cohesion. 
Based on the idea of social \textit{homophily}, users with many reciprocated ties are crucial in analysing socially cohesive groups. 
%
%\subsection{Utility of the MCT framework}
Figure~\ref{fig:Twitter-ecosystem} shows that communities on Twitter can be formed in many ways. 
A bi-modality approach differs across networks with respect to the depth of the features associated with the structural and textual modalities \cite{balasubramanyan2011block,leskovec2012learning,yang2013community}. 
Bi-modalities, e.g., network structure, features and attributes of nodes, lead to better and more interpretable community detection results. 
In Twitter, the structural component is not fully captured as it relies on directed connections. 
\textit{MCT} exploits the usability of features in the detection of a local community through the impact analyses of both modalities, especially the structural one. 
We have shown that a structural component is useful in community detection and has minimal practical requirements. 
\textit{MCT} offers a compact way to find and represent co-occurring users or user groups, allowing to explore local and global clustering requirements. 

\section{Conclusion}
\label{sec:conclusion}

Many natural networks exhibit a certain degree of organisation, in which node groups form tightly connected units called communities. 
Community detection allows to understand the network structure and extract useful information. 
%Within the social network's ecosystem, social interactions are continually evolving to support a myriad of objects to remain connected, leading to a high interest in detection tasks.
Detecting socially cohesive communities on Twitter is still challenging. 
While many methods have been proposed, they often discover disparate communities, likely to be socially unrelated. 
We observed that the topology of eccentric connections contributes to the detection of socially unrelated users and encourages the propagation of spurious content. 
Consequently, we propose a \textit{multilevel clustering technique (MCT)} to identify socially cohesive user groups, i.e.~\textit{microcosms}, on Twitter. 

The proposed \textit{MCT} framework, jointly modelling structural and intrinsic textual features, contributes toward a methodological paradigm for cohesive community detection in a dynamic and heterogeneous social media. 
This is important because until recently, community detection algorithms focused on single modality, e.g.~using node attributes or connectivity. 
Recent studies that combine information modalities are limited in capturing the nuances and intricate connection structure in platforms, such as Twitter. 
To improve the identification of socially cohesive communities, \textit{MCT} offers a scalable detection strategy.
% that is robust against noise in network data. The study presents the experimental results from the \textit{MCT} and evaluation benchmark models, illustrating the efficacy of the approach. The \textit{MCT framework} targets a similarity within a community of users using global and local information. 
The approach addresses the problem of structurally unrelated users, by adding a layer of social cohesion to existing community detection methods. 
In summary, \textit{MCT} contributes: 
(1) a systematic exposition of community detection or clustering algorithms, 
(2) an in-depth utilisation of the bi-modality for community detection, and
(3) detection of network communities at various levels. 

A note on the proposed method's complexity is in order here. When the network size is huge, it is challenging to authoritatively specify when a given community detection algorithm will converge. Thus, we rely on a single iteration to analyse the algorithm's complexity, which will provide insights to its future performance. Let us assume that the execution complexity of a basic parameterised algorithm is $f(C)$, then the term $O(f(C) \times s \times r \times m)$, where \textit{s} is the number of comparisons in deciding the next cluster, \textit{r} is the size size and \textit{m} the number runs. Execution wise, the complexity of the algorithm is relatively low. However, it tends to increase with growing data-points, hence the needs for further improvement in future work.

%% file: pn-appendix.tex
\section*{Appendix A: Supplementary information}
\label{appendix}
\paragraph{Structural Communities: optimisation and interpretability}%\label{appendix:function-optimisation}
\subsubsection*{Structural Communities: optimisation and interpretability}
Recall that the \textit{network-communities} ($\mathcal{M}_{C_{ns}}^{n\times p}$) matrix is decomposed into its approximate constituents given by Eq.~\ref{eq:structural-matrix}, i.e.~$\mathcal{M}_{C_{ns}} \approx \mathcal{M}_{vr}\mathcal{M}_{C_{nr}}^T $ and the optimisation function (Eq.~\ref{eq:structural-optimisation}) given by: $min._{\mathcal{M}_{c_{vr}},\mathcal{M}_{c_{nr}}}||\mathcal{M}_{c_{ns}} - \mathcal{M}_{c_{vr}}\mathcal{M}_{c_{nr}}^T||_F^2$ subject to $\mathcal{M}_{c_{vr}},\mathcal{M}_{c_{nr}} \geq 0 $. %%\begin{equation}\label{eq:structural-optimisation} 
The following conventions are used to represent the matrices: $\mathcal{M}_{c_{ns}} \mapsto D$, $\mathcal{M}_{c_{vr}} \mapsto P=[p_{is}]$, $\mathcal{M}_{c_{nr}} \mapsto Q=[q_{js}]$. We follow the \textit{NMF} scheme \cite{lee1999learning} in the \textit{modelling of structural communities}.

\subsubsection*{Iterative Update} 
%The process of obtaining a satisfactory approximation of the factored matrix requires a training based on an iterative update, in which the object is to minimise the error margin between the original matrix ($D$) and its constituents ($P,Q$). 
%The act of maximising the retained energy (i.e.~ the Frobenius norm) is the same as minimising the loss defined by the sum of squares of the values in the matrix (compare with $||D||_F^2$), hence the sum of the Frobenius and should be equal to $||D||_F^2$ \citep{aggarwal2018machine}. To ensure flexible and liberal approach (with respect to the non-negative constraints in the NMF scheme (Eq.~\ref{eq:structural-optimisation})), we employ a \textit{Lagrangian relaxation} to optimise the parameters in Eq.~\ref{eq:structural-optimisation} by introducing a new set of parameters ($\alpha$ and $\beta$), known as the \textit{Lagrangian multipliers} to the corresponding entries of the optimisation parameters ($P,Q$). 
In response to the additional parameters ($\mathbf{\alpha},\mathbf{\beta}$ with values $\leq 0$,) induced by the \textit{Lagrangian relaxation}, the objective function $M_{sr}$ is given by the following equation: %(Eq.~\ref{eq:structural-main}): 
        $$M_{sr} = ||D-PQ^T||^2_F + \sum^n_{i=1}\sum^k_{s=1}p_{is}\alpha_{is} + \sum^d_{j=1}\sum^k_{s=1}q_{js}\beta_{js}$$ %\end{equation}
To solve the optimisation problem, the process begins with computing the gradient of the Lagrangian relaxation with respect to the first aspect of the \textit{minmax} (i.e.~minimisation) optimisation variables. %Although the introduction of $\mathbf{\alpha}$ and $\mathbf{\beta}$ offers a degree of flexibility (which comes with a cost), 
To achieve an optimal solution, the optimisation condition needs to be based on $P,Q$ only. Hence, to eliminate the introduced \textit{Lagrangian multipliers}, the \textit{KKT optimality condition}, which suggests that $p_{is}\alpha _{is} = 0$ and $q_{js}\beta _{js}=0$, is applied.  We then solve for the optimisation parameters as follows.
    \begin{equation}
        \label{eq:structural-part1}
        \begin{split}
        ||D-PQ^T||^2_F & = (D-PQ^T)^T(D-PQ) \\
        & = (D^T-P^TQ)(D-PQ) \\
        & = \underbrace{D^TD}_{1} - \underbrace{D^TPQ}_{2}-\underbrace{QP^TD}_{3}+\underbrace{QP^TPQ}_{4}
        \end{split}
    \end{equation} 
In Eq.~\ref{eq:structural-part1}, the second term (2) and third term (3) are equal and the fourth term (4) can be expressed in a quadratic form depending on the parameter of interest (for minimisation). Thus,
    \begin{equation}
        \label{eq:structural-Ppart}
        M_{sr} =  D^TD - 2QP^TD+P^2Q^TQ + \sum^n_{i=1}\sum^k_{s=1}p_{is}\alpha_{is} + \sum^d_{j=1}\sum^k_{s=1}q_{js}\beta_{js}
    \end{equation}
From Eq.~\ref{eq:structural-Ppart}, the partial differentiation with respect to $P$ gives:
    \begin{equation}
        \label{eq:structural-partial1}
        \begin{split}
             \frac{\partial}{\partial p_{is}} M_{sr}  & =  -(2DQ)_{is} + (2PQ^TQ)_{is} + \alpha _{is}\\
             & \mbox{divide by 2 and equate to zero}\\ 
             & = -(DQ)_{is} + (PQ^TQ)_{is} + \alpha _{is} = 0 \\
             & \mbox{to eliminate the relaxation parameters multiply with $p_{is}$ throughout}\\ 
             & = -(DQ)_{is}p_{is} + (PQ^TQ)_{is}p_{is} + \alpha _{is}p_{is} = 0 \\
             & \mbox{the term $\alpha _{is}p_{is}$ equates to 0 according to KKT optimality, thus} \\ 
             & (PQ^TQ)_{is}p_{is} = (DQ)_{is}p_{is} \\ 
             & \mbox{the update rule:} \\ 
             & p_{is} = \frac{(DQ)_{is}p_{is}}{(PQ^TQ)_{is}}
        \end{split}
    \end{equation}
The last term or expression in Eq.~\ref{eq:structural-partial1} is the update rule for the parameter $P$. A similar process applies to the parameter $Q$: %From Eq.~\ref{eq:structural-main}, we have: 
    \begin{equation}
        \label{eq:structural-Qpart}
        M_{sr} =  D^TD - 2QP^TD+P^TPQ^2 + \sum^n_{i=1}\sum^k_{s=1}p_{is}\alpha_{is} + \sum^d_{j=1}\sum^k_{s=1}q_{js}\beta_{js}
    \end{equation}
The partial derivative with respect to $Q$ is given by the following: 
    \begin{equation}
        \label{eq:structural-partial2}
        \begin{split}
             \frac{\partial }{\partial q_{js}} M_{sr}  & =  -(2D^TP)_{js} + (2QP^TP)_{js} + \beta _{js}\\
             & \mbox{divide by 2 and equate to zero}\\ 
             & = -(D^TP)_{js} + (QP^TP)_{js} + \beta _{js} = 0 \\
             & \mbox{to eliminate the relaxation parameters multiply with $q_{js}$ throughout}\\ 
             & = -(D^TP)_{js}q_{js} + (QP^TP)_{js}q_{js} + \beta _{js}q_{js} = 0 \\
             & \mbox{the term $\beta _{js}q_{js}$ equates to 0 according to KKT optimality, thus} \\ 
             & (QP^TP)_{js}q_{js} = (D^TP)_{js}q_{js} \\ 
             & \mbox{the update rule:} \\ 
             & q_{js} = \frac{(D^TP)_{js}q_{js}}{(QP^TP)_{js}}
        \end{split}
    \end{equation}
The last term or expression in Eq.~\ref{eq:structural-partial2} is the update rule for the parameter $Q$. The process of updating $P,Q$ involves comparing their values to the original matrix $D$, and the goal is to minimise the difference or \textit{error}. The iterative update of the parameters ($p_{is}$ and $q_{js}$) continues until convergence. %We use a two-way decomposition (e.g., $D=PQ^T$) such that $D \in R^{n\times p}$ denotes the the original matrix to decompose or factorise into the following components: $P \in R^{n\times k}$ and $P \in R^{k\times k}$ using the following generic framework: